\documentclass[10pt,twocolumn,letterpaper]{article}

\usepackage{iccv}
\usepackage{times}
\usepackage{epsfig}
\usepackage{graphicx}
\usepackage{amsmath}
\usepackage{amssymb}
\usepackage{microtype}

\usepackage{blindtext}

\usepackage[pagebackref=true,breaklinks=true,letterpaper=true,colorlinks,bookmarks=false]{hyperref}

\iccvfinalcopy 

\def\iccvPaperID{6962} 

\ificcvfinal\fi

\newcommand{\myvec}[1]{\ensuremath{\vec{#1}}} 

\usepackage{array}
\newcolumntype{x}[1]{>{\centering\arraybackslash\hspace{0pt}}p{#1}}

\usepackage[inline]{enumitem}

\usepackage{stackengine}
\newcommand{\insetwhite}[2]{%
    \stackinset{l}{2pt}{b}{2.5pt}{%
    \textsf{\color{white}\footnotesize#1}}{#2}%
}

\newcommand\width{1.0\linewidth}

\usepackage[numbers,sort,compress]{natbib}

\begin{document}

\title{Fake it till you make it: face analysis in the wild using synthetic data alone}

\author{
Erroll Wood\thanks{Denotes equal contribution.} \quad
Tadas Baltru\v{s}aitis\footnotemark[1] \quad
Charlie Hewitt \quad
Sebastian Dziadzio  \\
Matthew Johnson \quad
Virginia Estellers \quad
Thomas J. Cashman \quad
Jamie Shotton \\[0.5em]
Microsoft
}

\newcommand\blfootnote[1]{%
  \begingroup
  \renewcommand\thefootnote{}\footnote{#1}%
  \addtocounter{footnote}{-1}%
  \endgroup
}

\maketitle
\ificcvfinal\thispagestyle{empty}\fi

\begin{abstract}
We demonstrate that it is possible to perform face-related computer vision in the wild using synthetic data alone.
The community has long enjoyed the benefits of synthesizing training data with graphics, but the domain gap between real and synthetic data has remained a problem, especially for human faces.
Researchers have tried to bridge this gap with data mixing, domain adaptation, and domain-adversarial training, but we show that it is possible to synthesize data with minimal domain gap, so that models trained on synthetic data generalize to real in-the-wild datasets.
We describe how to combine a procedurally-generated parametric 3D face model with a comprehensive library of hand-crafted assets to render training images with unprecedented realism and diversity.
We train machine learning systems for face-related tasks such as landmark localization and face parsing,
showing that synthetic data can both match real data in accuracy as well as open up new approaches where manual labeling would be impossible.
\end{abstract}

\section{Introduction}

When faced with a machine learning problem, the hardest challenge often isn't choosing the right machine learning model, it's finding the right data.
\blfootnote{\url{https://microsoft.github.io/FaceSynthetics}}
This is especially difficult in the realm of human-related computer vision, where concerns about the fairness of models and the ethics of deployment are paramount~\cite{karkkainen2019fairface}.
Instead of collecting and labelling real data, which is slow, expensive, and subject to bias, it can be preferable to \emph{synthesize} training data using computer graphics~\cite{Wood2015}.
With synthetic data, you can guarantee perfect labels without annotation noise, generate rich labels that are otherwise impossible to label by hand, and have full control over variation and diversity in a dataset.

\begin{figure}
    \centering
    \renewcommand\width{0.333\linewidth}
    \includegraphics[width=\width]{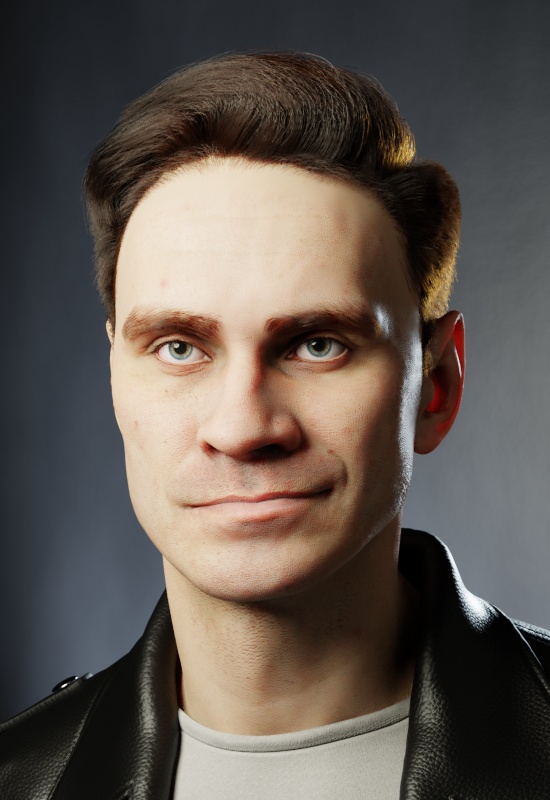}%
    \includegraphics[width=\width]{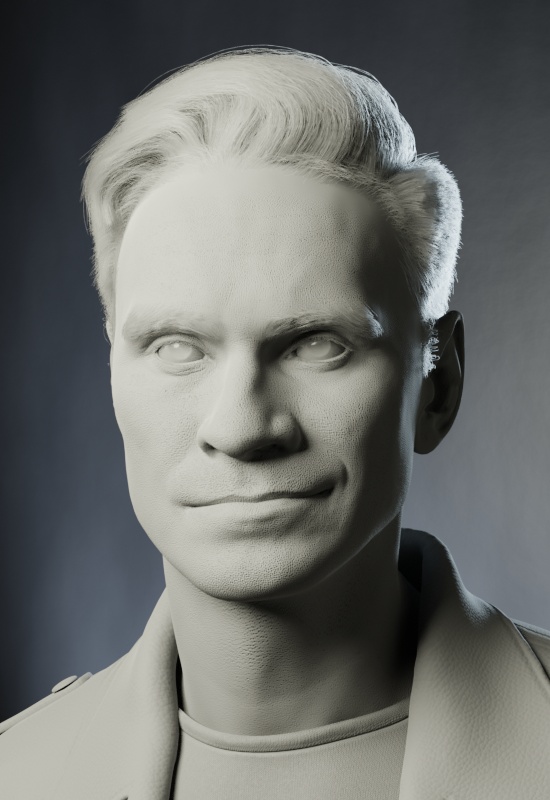}%
    \includegraphics[width=\width]{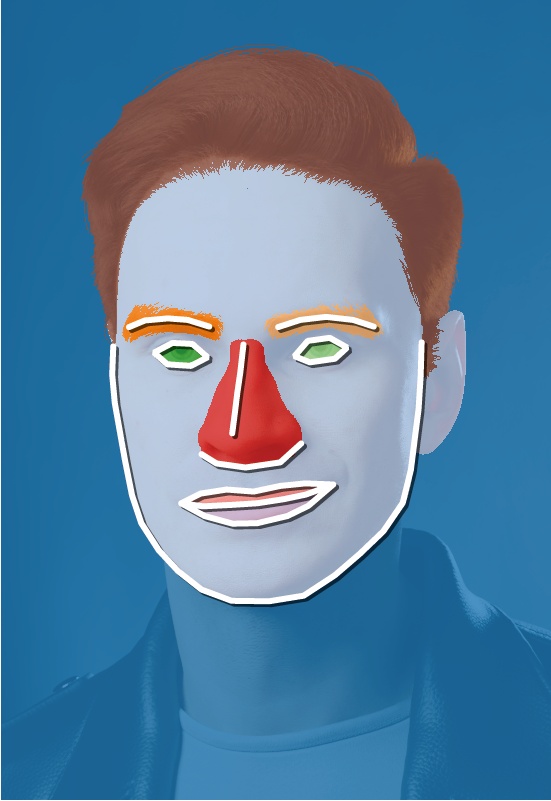}\\
    \vspace{0.1cm}
    \includegraphics[width=0.2\linewidth]{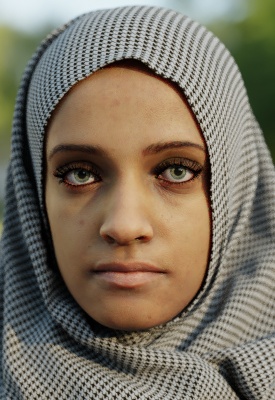}%
    \includegraphics[width=0.2\linewidth]{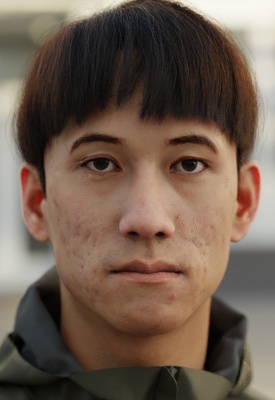}%
    \includegraphics[width=0.2\linewidth]{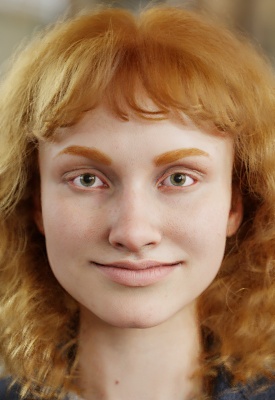}%
    \includegraphics[width=0.2\linewidth]{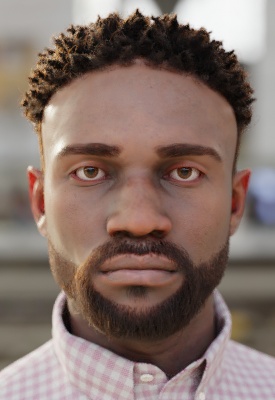}%
    \includegraphics[width=0.2\linewidth]{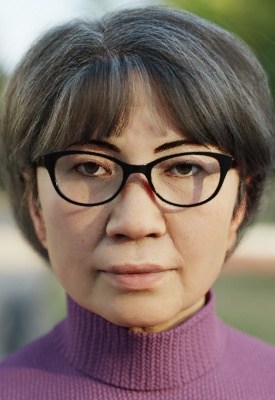}
    \caption{We render training images of faces with unprecedented realism and diversity. The first example above is shown along with 3D geometry and accompanying labels for machine learning.}
    \label{fig:intro_figure}
\end{figure}

\newcommand\subfigwidth{0.138\linewidth}

\begin{figure*}
    \includegraphics[width=\subfigwidth]{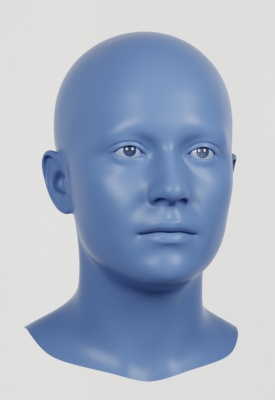} \hfill
    \includegraphics[width=\subfigwidth]{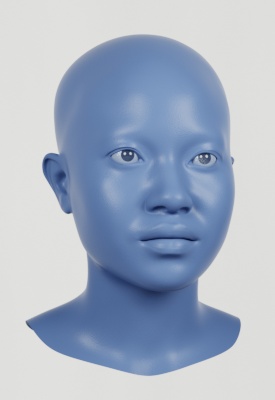} \hfill
    \includegraphics[width=\subfigwidth]{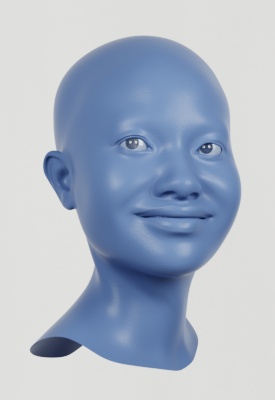} \hfill
    \includegraphics[width=\subfigwidth]{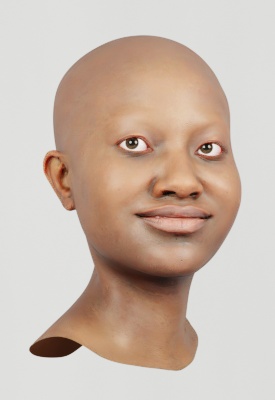} \hfill
    \includegraphics[width=\subfigwidth]{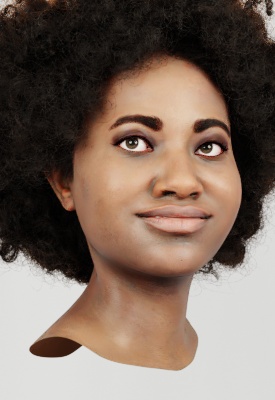} \hfill
    \includegraphics[width=\subfigwidth]{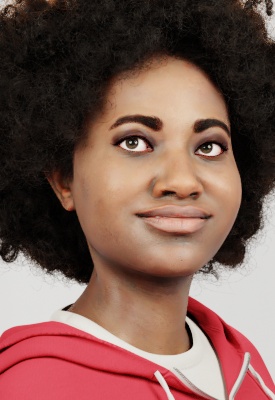} \hfill
    \includegraphics[width=\subfigwidth]{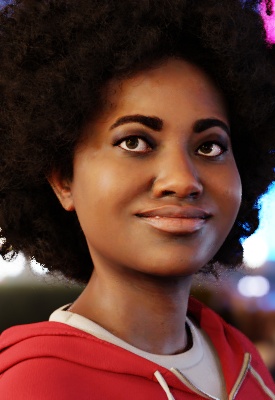}\\
    \makebox[\subfigwidth]{\small Template face}
    \makebox[\subfigwidth]{\small + identity}
    \makebox[\subfigwidth]{\small + expression}
    \makebox[\subfigwidth]{\small + texture}
    \makebox[\subfigwidth]{\small + hair}
    \makebox[\subfigwidth]{\small + clothes}
    \makebox[\subfigwidth]{\small + environment} \vspace{0.2em}
    \caption{We procedurally construct synthetic faces that are realistic and expressive. Starting with our template face, we randomize the identity, choose a random expression, apply a random texture, attach random hair and clothing, and render the face in a random environment.}
    \label{fig:method}
\end{figure*}

Rendering convincing humans is one of the hardest problems in computer graphics.
Movies and video games have shown that realistic digital humans are possible, but with significant artist effort per individual~\cite{senua2016,Avengers2018}.
While it's possible to generate endless novel face images with recent self-supervised approaches~\cite{karras2020analyzing}, corresponding labels for supervised learning are not available.
As a result, previous work has resorted to synthesizing facial training data with simplifications, with results that are far from realistic.
We have seen progress in efforts that attempt to cross the domain gap using domain adaptation~\cite{shrivastava2017learning} by refining synthetic images to look more real, and domain-adversarial training~\cite{ganin2016advers} where machine learning models are encouraged to ignore differences between the synthetic and real domains, but less work has attempted to improve the quality of synthetic data itself.
Synthesizing realistic face data has been considered so hard that we encounter the assumption that synthetic data cannot fully replace real data for problems in the wild~\cite{shrivastava2017learning}.

In this paper we demonstrate that the opportunities for synthetic data are much wider than previously realised, and are achievable today.
We present a new method of acquiring training data for faces -- rendering 3D face models with an unprecedented level of realism and diversity (see \autoref{fig:intro_figure}).
With a sufficiently good synthetic framework, it is possible to create training data that can be used to solve real world problems in the wild, without using any real data at all.

It requires considerable expertise and investment to develop a synthetics framework with minimal domain gap.
However, once implemented, it becomes possible to generate a wide variety of  training data with minimal incremental effort.
Let's consider some examples; say you have spent time labelling face images with landmarks.
However, you suddenly require additional landmarks in each image.
Relabelling and verifying will take a long time, but
with synthetics, you can regenerate clean and consistent labels at a moment's notice.
Or, say you are developing computer vision algorithms for a new camera, e.g.\ an infrared face-recognition camera in a mobile phone.
Few, if any, hardware prototypes may exist, making it hard to collect a dataset.
Synthetics lets you render faces from a simulated device to develop algorithms and even guide hardware design itself.

We synthesize face images by procedurally combining a parametric face model with a large library of high-quality artist-created assets, including textures, hair, and clothing (see \autoref{fig:method}).
With this data we train models for common face-related tasks: face parsing and landmark localization.
Our experiments show that models trained with a single generic synthetic dataset can be just as accurate as those trained with task-specific real datasets, achieving results in line with the state of the art.
This opens the door to other face-related tasks that can be confidently addressed with synthetic data instead of real.

Our contributions are as follows.
First, we describe how to synthesize realistic and diverse training data for face analysis in the wild, achieving results in line with the state of the art.
Second, we present ablation studies that validate the steps taken to achieve photorealism.
Third is the synthetic dataset itself, which is available from our project webpage: {\small \url{https://microsoft.github.io/FaceSynthetics}}.

\section{Related work}

Diverse face datasets are very difficult to collect and annotate.
Collection techniques such as web crawling pose significant privacy and copyright concerns.
Manual annotation is error-prone and can often result in inconsistent labels.
Hence, the research community is increasingly looking at augmenting or replacing real data with synthetic.

\subsection{Synthetic face data}

The computer vision community has used synthetic data for many tasks, including
object recognition~\cite{Rozantsev2014, qiu2017unrealcv, hodavn2019photorealistic, Yao2020},
scene understanding~\cite{richter2016playing, Gaidon2016, ros2016, kar2019metasim},
eye tracking~\cite{Swirski2014, Wood2015},
hand tracking~\cite{simon2017hand,GANeratedHands_CVPR2018}, and
full-body analysis~\cite{shotton2011real, varol17_surreal, Ning2003}.
However, relatively little previous work has attempted to generate full-face synthetics using computer graphics, due to the complexity of modeling the human head.

A common approach is to use a 3D Morphable Model (3DMM)~\cite{blanz1999morphable}, since these can provide consistent labels for different faces.
Previous work has focused on parts of the face such as the eye region~\cite{Sugano2014} or the \emph{hockey mask}~\cite{Zeng2019, Richardson2016}.
\citet{Zeng2019}, \citet{Richardson2017}, and \citet{Sela2017} used 3DMMs to render training data for reconstructing detailed facial geometry.
Similarly, \citet{Wood2016} rendered an eye region 3DMM for gaze estimation. 
However, since these approaches only render part of the face, the resulting data has limited use for tasks that consider the whole face.

Building parametric models is challenging, so an alternative is to render 3D scans directly~\cite{Baltrusaitis2012, Sugano2014, Wood2015, saito20pifu}. 
\citet{Jeni2015} rendered the BU-4DFE dataset~\cite{BU4DFE} for dense 3D face alignment,
and \citet{kuhnke2019deep} rendered commercially-available 3D head scans for head pose estimation.
While often realistic, these approaches are limited by the diversity expressed in the scans themselves, and cannot provide rich semantic labels for machine learning.

Manipulating 2D images can be an alternative to using a 3D graphics pipeline.
\citet{Zhu2017} fit a 3DMM to face images, and warped them to augment the head pose.
\citet{Nojavanasghari2017} composited hand images onto faces to improve face detection.
These approaches can only make minor adjustments to existing images, limiting their use.

\subsection{Training with synthetic data}

Although it is common to rely on synthetic data alone for full-body tasks~\cite{shotton2011real, saito2019pifu}, synthetic data is rarely used on its own for face-related machine learning.
Instead it is either first adapted to make it look more like some target domain, or used alongside real data for pre-training~\cite{Zeng2019} or regularizing models~\cite{KowalskiECCV2020,Gecer2018}.
The reason for this is the \emph{domain gap} -- a difference in distributions between real and synthetic data which makes generalization difficult~\cite{kar2019metasim}.

Learned domain adaptation modifies synthetic images to better match the appearance of real images.
\citet{shrivastava2017learning} use an adversarial refiner network to adapt synthetic eye images with regularization to preserve annotations.
Similarly, \citet{bak2018domain} adapt synthetic data using a CycleGAN~\cite{CycleGAN2017} with a regularization term for preserving identities. 
A limitation of learned domain adaptation is the tendency for image semantics to change during adaptation~\cite{garbin2020high}, hence the need for regularization~\cite{bak2018domain, shrivastava2017learning, GANeratedHands_CVPR2018}.
These techniques are therefore unsuitable for fine-grained annotations, such as per-pixel labels or precise landmark coordinates.


Instead of adapting data, it is possible to learn features that are resistant to the differences between domains~\cite{ganin2016advers, sankaranarayanan2018learning}.
\citet{wu2020synthetictoreal} mix real and synthetic data through a domain classifier to learn domain-invariant features for text detection, and
\citet{saleh2018effective} exploit the observation that shape is less affected by the domain gap than appearance for scene semantic segmentation.

In our work, we do not perform any of these techniques and instead minimize the domain gap at the source, by generating highly realistic synthetic data.

\section{Synthesizing face images}

The Visual Effects (VFX) industry has developed many techniques for convincing audiences that 3D faces are real, and we build upon these in our approach.
However, a key difference is scale: while VFX might be used for a handful of actors, we require diverse training data of thousands of synthetic individuals.
To address this, we use procedural generation to randomly create and render novel 3D faces without any manual intervention.

We start by sampling a generative 3D face model that captures the diversity of the human population.
We then randomly `dress up' each face with samples from large collections of hair, clothing, and accessory assets.
All collections are sampled independently to create synthetic individuals who are as diverse as possible from one another.
This section describes the technical components we built in order to enable asset collections that can be mixed-and-matched atop 3D faces in a random, yet plausible manner.

\subsection{3D face model}

\begin{figure}
    \includegraphics[width=\linewidth]{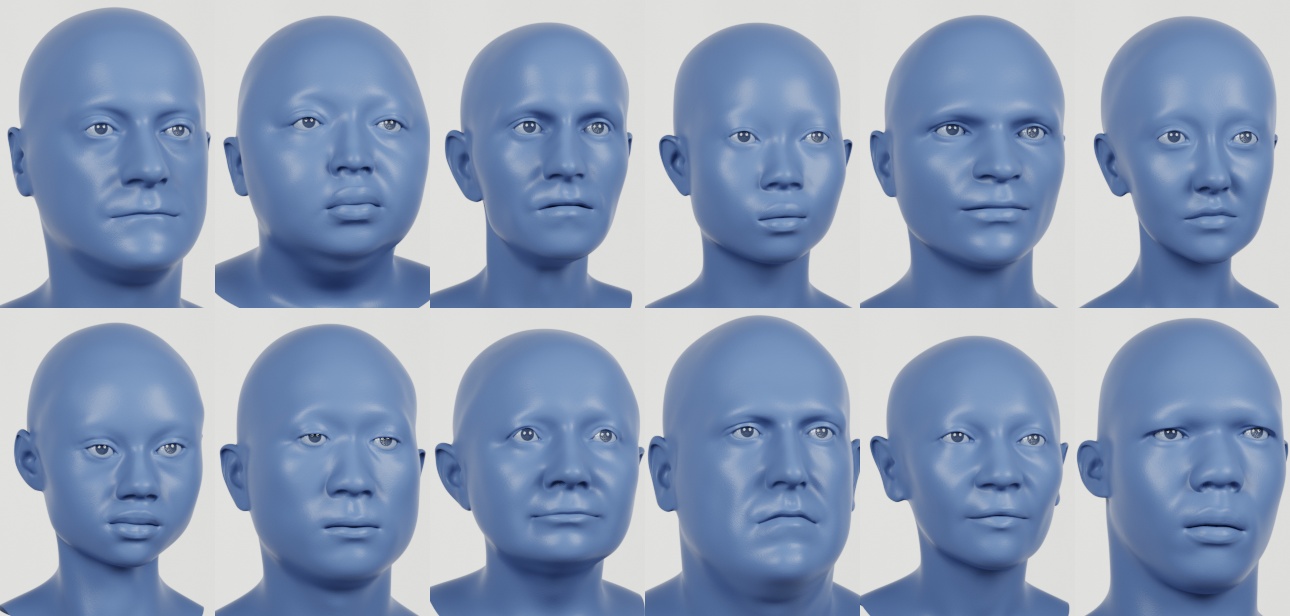}
    \caption{3D faces sampled from our generative model, demonstrating how our model captures the diversity of the human population.}
    \label{fig:3d_face_model_samples}
\end{figure}

\begin{figure}
    \includegraphics[width=\linewidth]{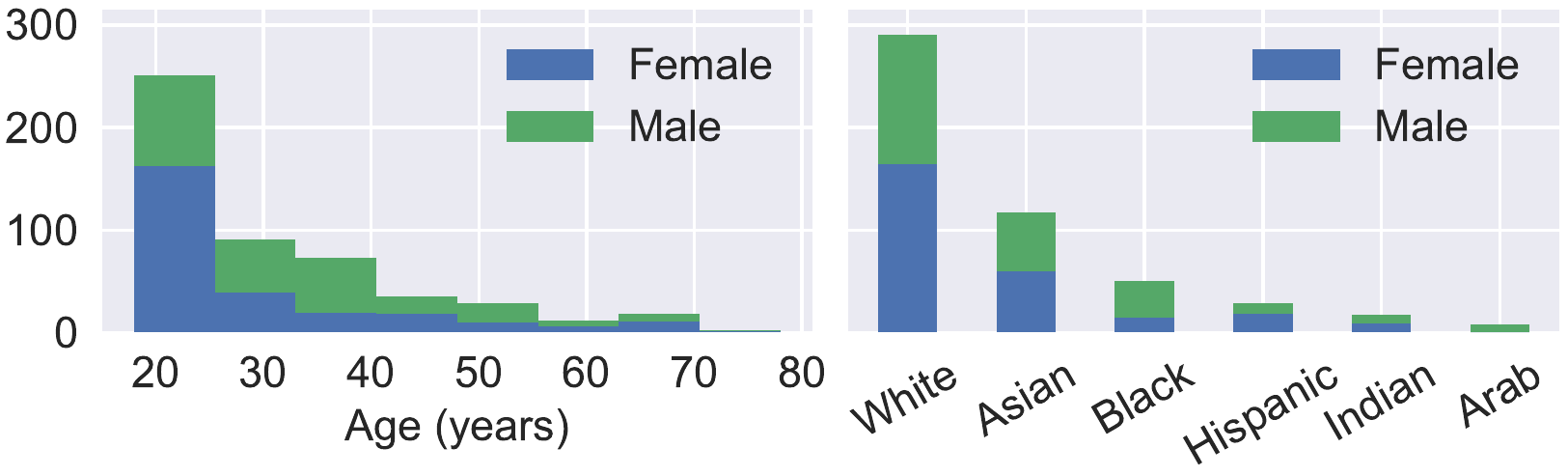}
    \caption{Histograms of self-reported age, gender, and ethnicity in our scan collection, which was used to build our face model and texture library. Our collection covers a range of age and ethnicity.}
    \label{fig:3d_face_model_charts}
\end{figure}

Our generative 3D face model captures how face shape varies across the human population, and changes during facial expressions.
It is a blendshape-based face rig similar to previous work~\cite{Li2017flame,Gerig2017}, and comprises a mesh of $N\!=\!7,667$ vertices and $7,414$ polygons, and a minimal skeleton of $K\!=\!4$ joints: the head, neck, and two eyes.

The face mesh vertex positions are defined by mesh generating function
$\mathcal{M}(\myvec{\beta}, \myvec{\psi}, \myvec{\theta})\!:\!\mathbb{R}^{|\myvec{\beta}|\times|\myvec{\psi}|\times|\myvec{\theta}|}\!\to\!\mathbb{R}^{N\times3}$ which takes parameters
$\myvec{\beta}\in\mathbb{R}^{|\myvec{\beta}|}$ for identity,
$\myvec{\psi}\in\mathbb{R}^{|\myvec{\psi}|}$ for expression, and
$\myvec{\theta}\in\mathbb{R}^{K\times3}$ for skeletal pose.
The pose parameters~$\myvec{\theta}$ are per-joint local rotations represented as Euler angles. $\mathcal{M}$~is defined as
$$
\mathcal{M}(\myvec{\beta}, \myvec{\psi}, \myvec{\theta}) =
\mathcal{L}(\mathcal{T}(\myvec{\beta}, \myvec{\psi}), \myvec{\theta}, \mathcal{J}(\myvec{\beta}); \mathbf{W})
$$
where $\mathcal{L}(\mathbf{X}, \myvec{\theta}, \mathbf{J}; \mathbf{W})$ is a standard linear blend skinning (LBS) function~\cite{Lewis2000skinning} that rotates vertex positions $\mathbf{X}\in\mathbb{R}^{N\times3}$ about joint locations $\mathbf{J}\in\mathbb{R}^{K\times3}$ by local joint rotations \myvec{\theta}, with per-vertex weights $\mathbf{W}\in\mathbb{R}^{K\times N}$ determining how rotations are 
interpolated across the mesh.
$\mathcal{T}(\myvec{\beta}, \myvec{\psi})\!:\!\mathbb{R}^{|\myvec{\beta}| \times |\myvec{\psi}|}\to\mathbb{R}^{N\times3}$
constructs a face mesh in the bind pose by adding displacements to the template mesh $\mathbf{\overline{T}}\!\in\!\mathbb{R}^{N \times 3}$, which represents the average face with neutral expression:
$$
\mathcal{T}(\myvec{\beta}, \myvec{\psi})^j_{\:k} =
\overline{T}^j_{\:k} +
\beta_i S^{ij}_{\;\;k} +
\psi_i E^{ij}_{\;\;k}
$$
given linear identity basis $\mathbf{S}\!\in\!\mathbb{R}^{|\myvec{\beta}| \times N \times 3}$ and
expression basis $\mathbf{E}\!\in\!\mathbb{R}^{|\myvec{\psi}| \times N \times 3}$.
Note the use of Einstein summation notation in this definition and below.
Finally, $\mathcal{J}(\myvec{\beta})\!:\!\mathbb{R}^{|\myvec{\beta}|}\to\mathbb{R}^{K\times3}$ moves the template joint locations $\mathbf{\overline{J}}\!\in\!\mathbb{R}^{K \times 3}$ to account for changes in identity:
$$
\mathcal{J}(\myvec{\beta})^j_{\:k} =
\overline{J}^j_{\:k} +
W^j_{\;l} \beta_i S^{il}_{\;\;k}.
$$

\begin{figure}

    \insetwhite{Raw}{\includegraphics[width=0.164\linewidth]{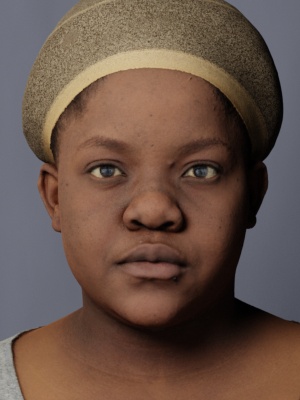}}%
    \insetwhite{Clean}{\includegraphics[width=0.164\linewidth]{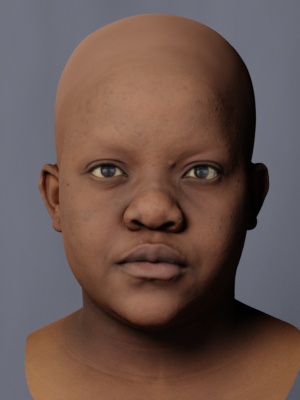}}\hfill
    \includegraphics[width=0.164\linewidth]{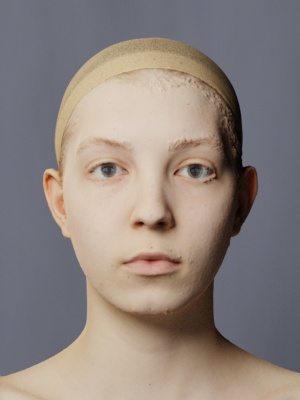}%
    \includegraphics[width=0.164\linewidth]{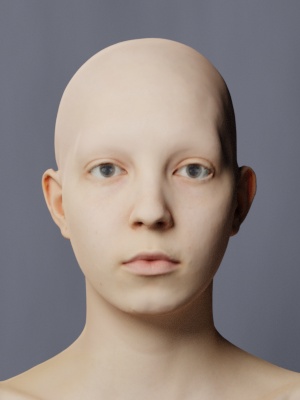}\hfill
    \includegraphics[width=0.164\linewidth]{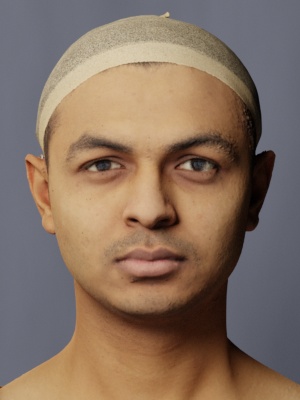}%
    \includegraphics[width=0.164\linewidth]{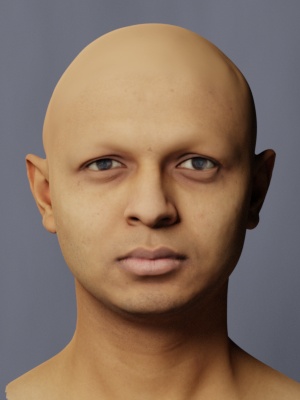}%
    \caption{We manually ``clean'' raw high-resolution 3D head scans to remove noise and hair. We use the resulting clean scans to build our generative geometry model and texture library.}
    \label{fig:cleaned_scans}
\end{figure}

We learn the identity basis $\mathbf{S}$ from high quality 3D scans of $M\!=\!511$ individuals with neutral expression.
Each scan was cleaned (see \autoref{fig:cleaned_scans}), and registered to the topology of $\mathbf{\overline{T}}$ using commercial software~\cite{Wrap3}, resulting in training dataset $\mathbf{V}\in\mathbb{R}^{M \times 3N}$.
We then jointly fit identity basis $\mathbf{S}$ and parameters $[\myvec{\beta}_1, \ldots, \myvec{\beta}_M]$ to $\mathbf{V}$.
In order to generate novel face shapes, we fit a multivariate normal distribution to the fitted identity parameters, and sample from it (see \autoref{fig:3d_face_model_samples}).
As is common in computer animation, both expression basis $\mathbf{E}$ and skinning weights $\mathbf{W}$ were authored by an artist, and are kept fixed while learning $\mathbf{S}$.

\subsection{Expression}

\begin{figure}
    \includegraphics[width=\linewidth]{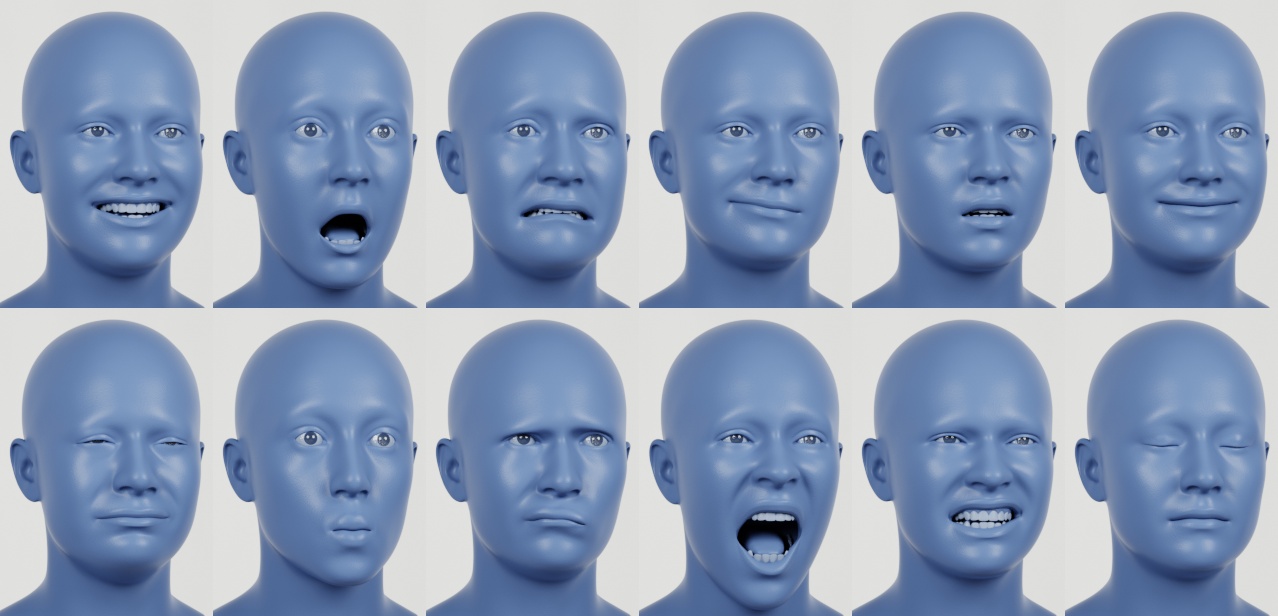}
    \caption{Examples from our data-driven expression library and manually animated sequence, visualized on our template face.}
    \label{fig:expression_examples}
\end{figure}

\newcommand{\expressionlibrarysize}{27,000}
\newcommand{\expressionlibrarynumdims}{54}

We apply random expressions to each face so that our downstream machine learning models are robust to facial motion.
We use two sources of facial expression.
Our primary source is a library of \expressionlibrarysize{} expression parameters $\{\myvec{\psi}_i\}$ built by fitting a 3D face model to a corpus of 2D images with annotated face landmarks.
However, since the annotated landmarks are sparse, it is not possible to recover all types of expression from these landmarks alone, e.g.\ cheek puffs.
Therefore, we additionally sample expressions from a manually animated sequence that was designed to fill the gaps in our expression library by exercising the face in realistic, but extreme ways.
\autoref{fig:expression_examples} shows samples from our expression collection. 
In addition to facial expression, we layer random eye gaze directions on top of sampled expressions, and use procedural logic to pose the eyelids accordingly.

\subsection{Texture}

\newcommand{\numtextures}{$200$}

\begin{figure}
    \renewcommand\width{0.24\linewidth}
    \insetwhite{Face model\vphantom{p}}{\includegraphics[width=\width]{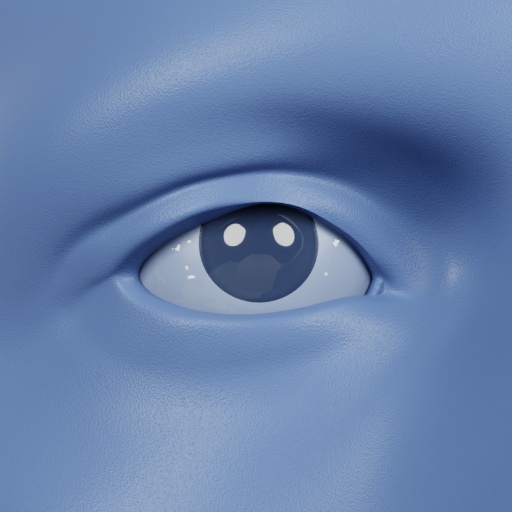}}\hfill
    \insetwhite{+coarse disp.}{\includegraphics[width=\width]{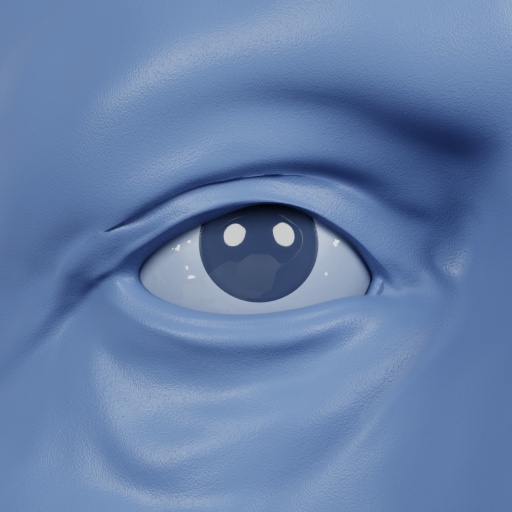}}\hfill
    \insetwhite{+meso disp.}{\includegraphics[width=\width]{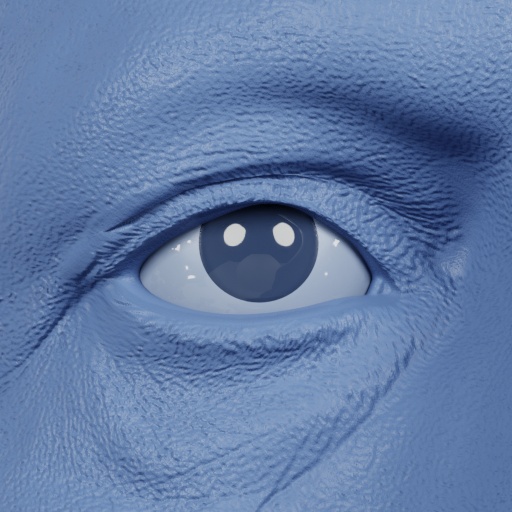}}\hfill
    \includegraphics[width=\width]{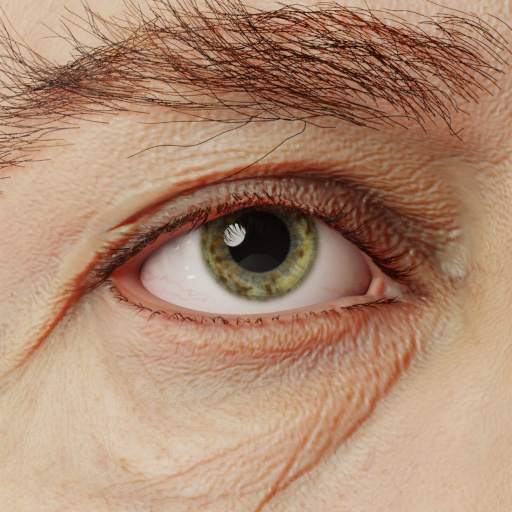}
    \caption{We apply coarse and meso-displacement to our 3D face model to ensure faces look realistic even when viewed close-up.}
    \label{fig:dispmaps}
\end{figure}

Synthetic faces should look realistic even when viewed at extremely close range, for example by an eye-tracking camera in a head-mounted device.
To achieve this, we collected \numtextures{} sets of high resolution (8192$\!\times\!$8192 px) textures from our cleaned face scans.
For each scan, we extract one albedo texture for skin color, and two displacement maps (see \autoref{fig:dispmaps}).
The coarse displacement map encodes scan geometry that is not captured by the sparse nature of our vertex-level identity model.
The meso-displacement map approximates skin-pore level detail and is built by high-pass filtering the albedo texture, assuming that dark pixels correspond to slightly recessed parts of the skin.

Unlike previous work~\cite{Zeng2019, Richardson2016}, we do not build a generative model of texture, as such models struggle to faithfully produce high-frequency details like wrinkles and pores.
Instead, we simply pick a corresponding set of albedo and displacement textures from each scan.
The textures are combined in a physically-based skin material featuring subsurface scattering~\cite{Christensen2015}. 
Finally, we optionally apply makeup effects to simulate eyeshadow, eyeliner and mascara.


\subsection{Hair}

\newcommand{\numheadhairs}{512}
\newcommand{\numeyebrows}{162}
\newcommand{\numbeards}{142}
\newcommand{\numeyelashes}{42}

\begin{figure}
    \renewcommand{\arraystretch}{0}
    \begin{tabular}{@{}c@{}c@{}c@{}c@{}c@{}}
        \includegraphics[width=0.2\linewidth]{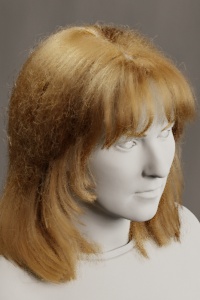}&%
        \includegraphics[width=0.2\linewidth]{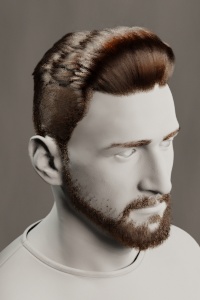}&%
        \includegraphics[width=0.2\linewidth]{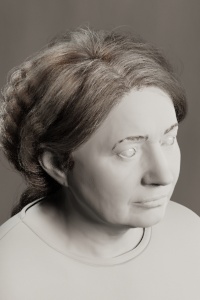}&%
        \includegraphics[width=0.2\linewidth]{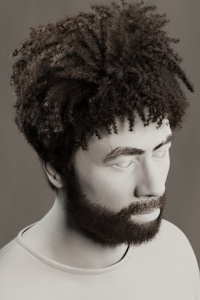}&%
        \includegraphics[width=0.2\linewidth]{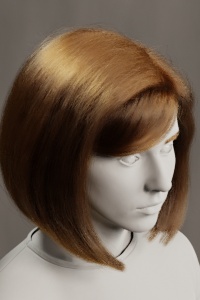}\\
        \includegraphics[width=0.2\linewidth]{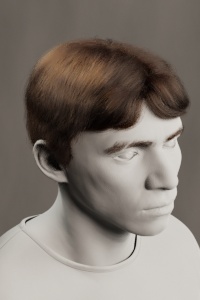}&%
        \includegraphics[width=0.2\linewidth]{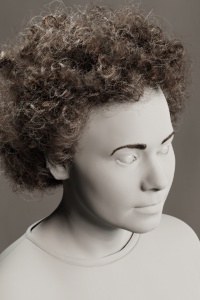}&%
        \includegraphics[width=0.2\linewidth]{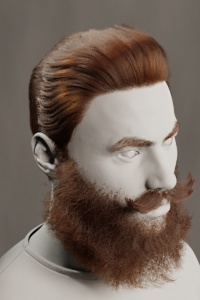}&%
        \includegraphics[width=0.2\linewidth]{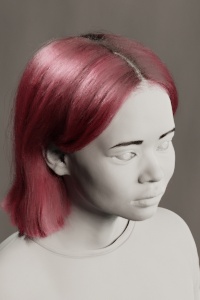}&%
        \includegraphics[width=0.2\linewidth]{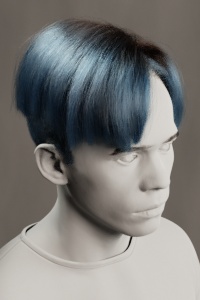}\\
    \end{tabular}\vspace{0.5em}

    \caption{Our hair library contains a diverse range of scalp hair, eyebrows, and beards. When assembling a 3D face, we choose hair style and appearance at random.}
    \label{fig:hairlibrary}
\end{figure}

In contrast to other work which approximates hair with textures or coarse geometry~\cite{saito20pifu, Gerig2017}, we represent hair as individual 3D strands, with a full head of hair comprising over 100,000 strands.
Modelling hair at the strand level allows us to capture realistic multi-path illumination effects.
Shown in \autoref{fig:hairlibrary}, our hair library includes \numheadhairs{} scalp hair styles, \numeyebrows{} eyebrows, \numbeards{} beards, and \numeyelashes{} sets of eyelashes.
Each asset was authored by a groom artist who specializes in creating digital hair.
At render time, we randomly combine scalp, eyebrow, beard, and eyelash grooms.

We use a physically-based procedural hair shader to accurately model the complex material properties of hair~\cite{chiang2016practical}.
This shader allows us to control the color of the hair with parameters for melanin~\cite{lozano2017diversity} and grayness, and even lets us dye or bleach the hair for less common hair styles.

\subsection{Clothing}

\begin{figure}
    \renewcommand{\arraystretch}{0}
    \renewcommand\width{0.199\linewidth}
    \begin{tabular}{@{}c@{}c@{}c@{}c@{}c@{}}
        \includegraphics[width=\width]{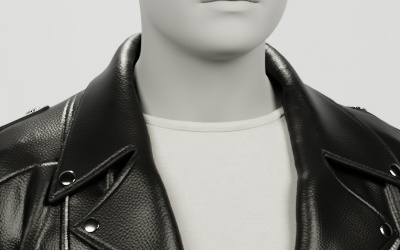}&%
        \includegraphics[width=\width]{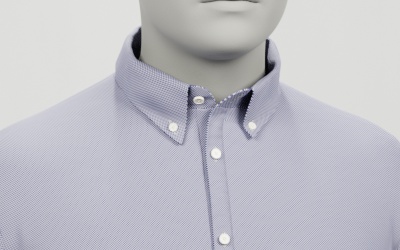}&%
        \includegraphics[width=\width]{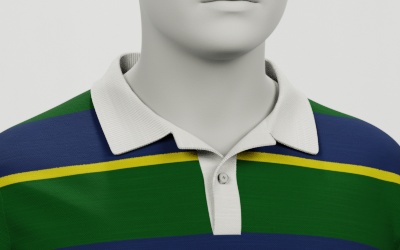}&%
        \includegraphics[width=\width]{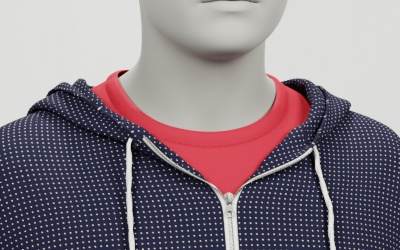}&%
        \includegraphics[width=\width]{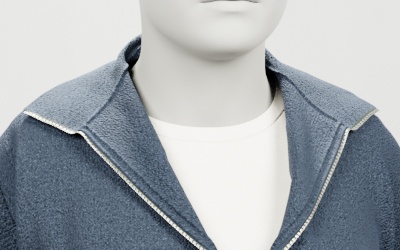}\\
        \includegraphics[width=\width]{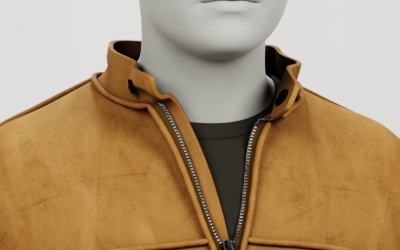}&%
        \includegraphics[width=\width]{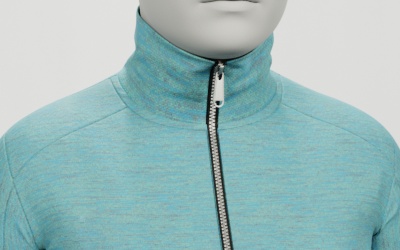}&%
        \includegraphics[width=\width]{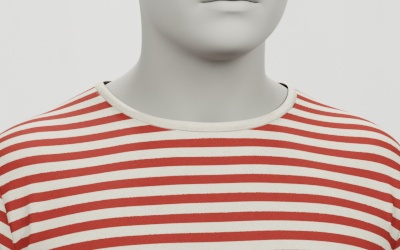}&%
        \includegraphics[width=\width]{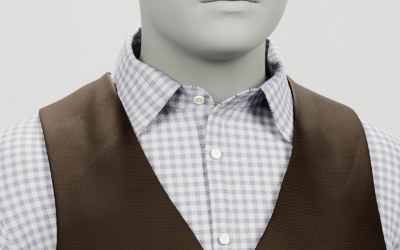}&%
        \includegraphics[width=\width]{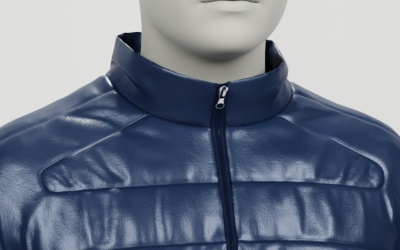}\\
        \includegraphics[width=\width]{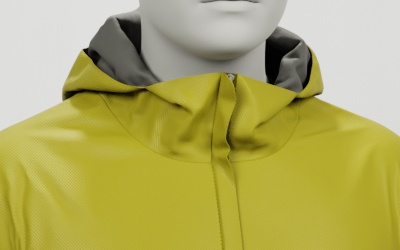}&%
        \includegraphics[width=\width]{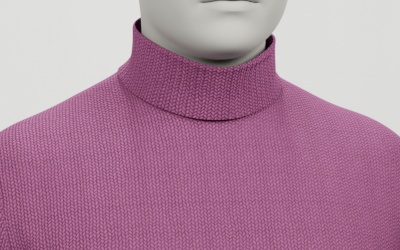}&%
        \includegraphics[width=\width]{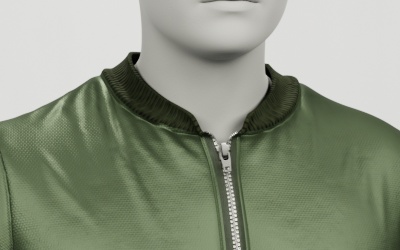}&%
        \includegraphics[width=\width]{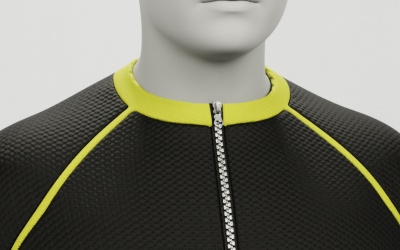}&%
        \includegraphics[width=\width]{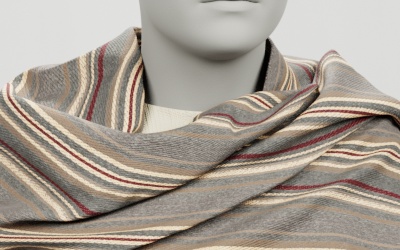}\\
    \end{tabular}\vspace{0.2em}

    \renewcommand\width{0.166\linewidth}
    \begin{tabular}{@{}c@{}c@{}c@{}c@{}c@{}c@{}}
        \includegraphics[width=\width]{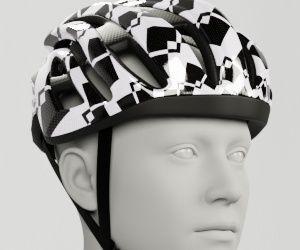}&%
        \includegraphics[width=\width]{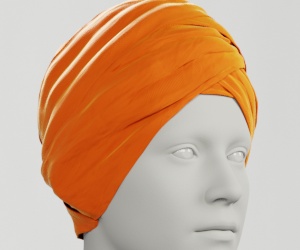}&%
        \includegraphics[width=\width]{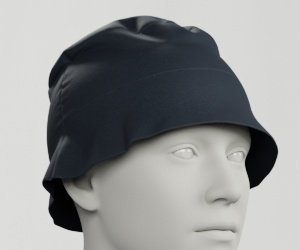}&%
        \includegraphics[width=\width]{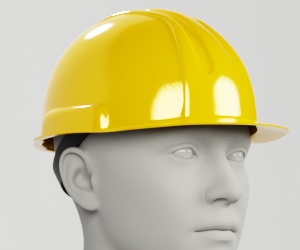}&%
        \includegraphics[width=\width]{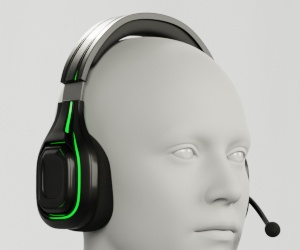}&%
        \includegraphics[width=\width]{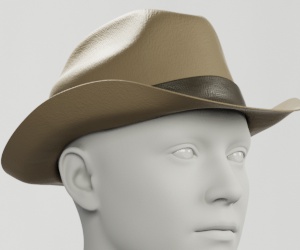}\\
    \end{tabular}\vspace{0.2em}

    \begin{tabular}{@{}c@{}c@{}c@{}c@{}c@{}c@{}}
        \includegraphics[width=\width]{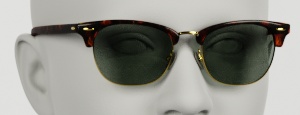}&%
        \includegraphics[width=\width]{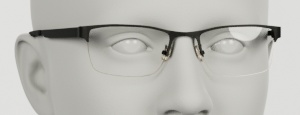}&%
        \includegraphics[width=\width]{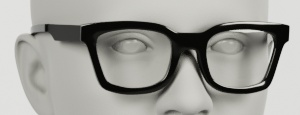}&%
        \includegraphics[width=\width]{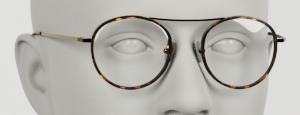}&%
        \includegraphics[width=\width]{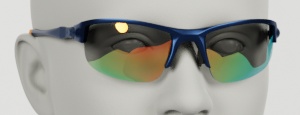}&%
        \includegraphics[width=\width]{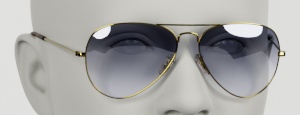}\\
    \end{tabular}\vspace{0.5em}

    \caption{Each face is dressed in a random outfit assembled from our digital wardrobe -- a collection of diverse 3D clothing and accessory assets that can be fit around our 3D head model.} 
    \label{fig:digitalwardrobe}
\end{figure}

\newcommand{\numacctops}{$30$}
\newcommand{\numacchead}{$36$}
\newcommand{\numaccface}{$7$}
\newcommand{\numacceye}{$11$}

Images of faces often include what someone is wearing, so we dress our faces in 3D clothing.
Our digital wardrobe contains \numacctops{} upper-body outfits which were manually created using clothing design and simulation software~\cite{MarvelousDesigner}.
As shown in \autoref{fig:digitalwardrobe}, these outfits include formal, casual, and athletic clothing.
In addition to upper-body garments, we dress our faces in headwear (\numacchead{} items), facewear (\numaccface{} items) and eyewear (\numacceye{} items) including helmets, head scarves, face masks, and eyeglasses.
All clothing items were authored on an unclothed body mesh with either the average male or female body proportions~\cite{SMPL:2015} in a relaxed stance.


We deform garments with a non-rigid cage-based deformation technique~\cite{Anderson2012parametric} so they fit snugly around different shaped faces.
Eyeglasses are rigged with a skeleton, and posed using inverse kinematics so the temples and nose-bridge rest on the corresponding parts of the face.

\subsection{Rendering}

\newcommand{\numHDRI}{448}

\begin{figure}
    \renewcommand\width{0.142\linewidth}
    \includegraphics[width=\width]{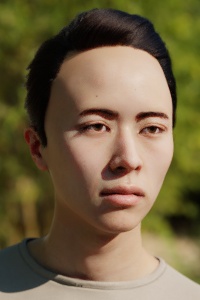}%
    \includegraphics[width=\width]{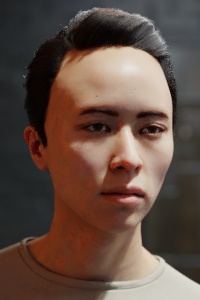}%
    \includegraphics[width=\width]{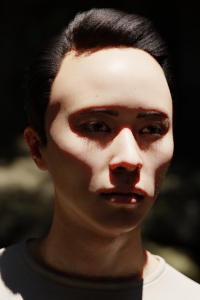}%
    \includegraphics[width=\width]{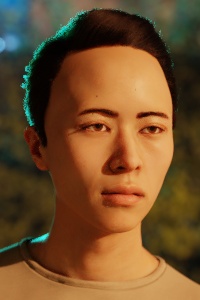}%
    \includegraphics[width=\width]{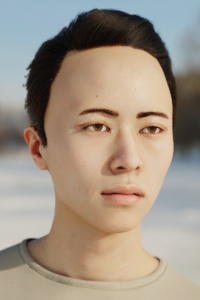}%
    \includegraphics[width=\width]{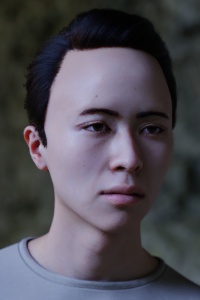}%
    \includegraphics[width=\width]{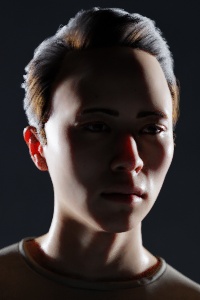}

    \caption{We use HDRIs to illuminate the face. The same face can look very different under different illumination.}
    \label{fig:hdris}
\end{figure}

\begin{figure}
    \centering
    \includegraphics[width=\width]{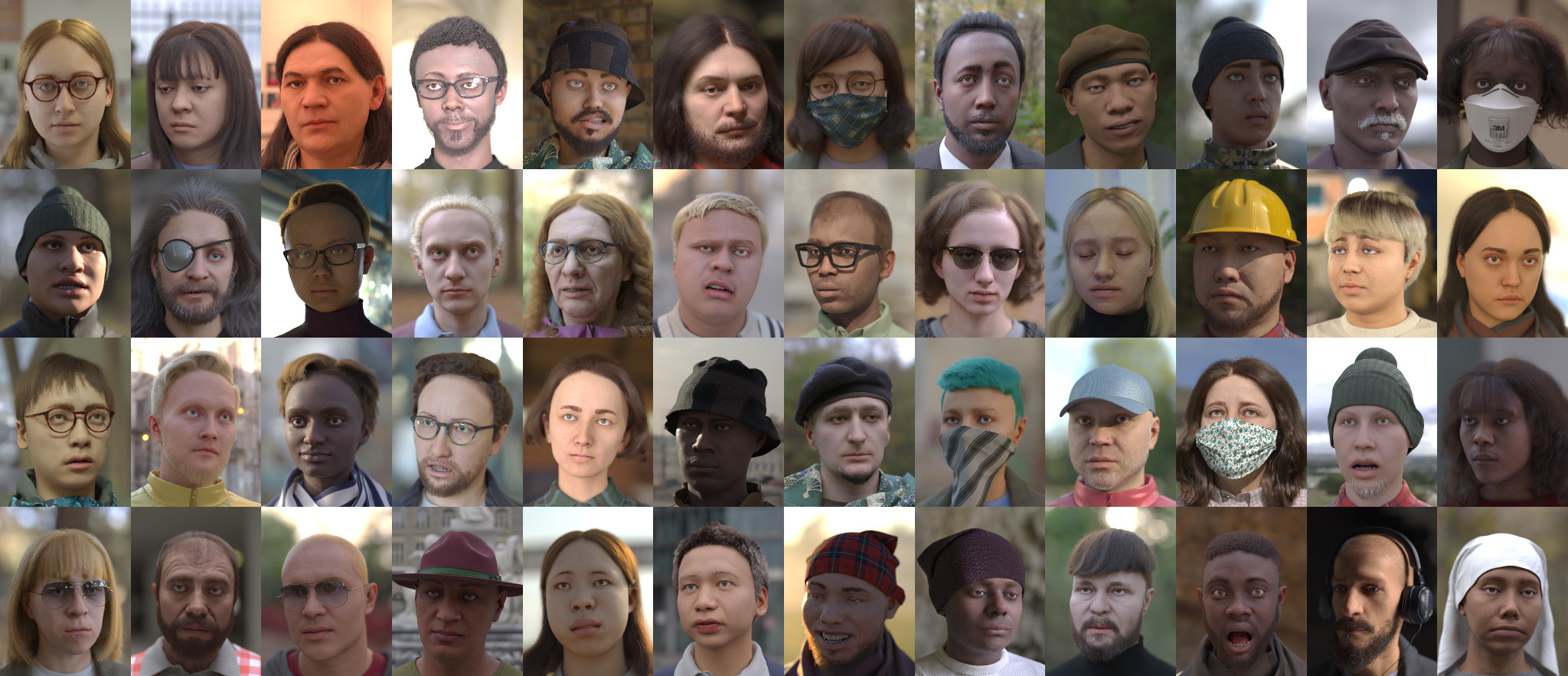}
    \caption{Examples of synthetic faces that we randomly generated and rendered for use as training data.}
    \label{fig:many_samples}
\end{figure}

We render face images with Cycles, a photorealistic ray-tracing renderer~\cite{CyclesRenderer}.
We randomly position a camera around the head, and point it towards the face.
The focal length and depth of field are varied to simulate different cameras and lenses.
We employ image-based lighting~\cite{debevec2006image} with high dynamic range images (HDRI) to illuminate the face and provide a background (see \autoref{fig:hdris}).
For each image, we randomly pick from a collection of \numHDRI{} HDRIs that include a range of different environments~\cite{HDRIHaven}.
See \autoref{fig:many_samples} for examples of faces rendered with our framework.

\begin{figure}
    \renewcommand\width{0.1666\linewidth}
    \insetwhite{Albedo\vphantom{p}}{\includegraphics[width=\width]{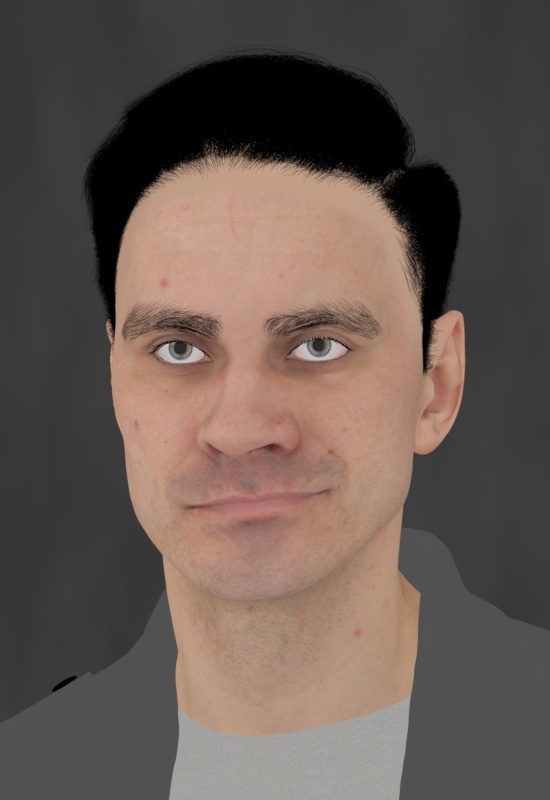}}%
    \insetwhite{Normals\vphantom{p}}{\includegraphics[width=\width]{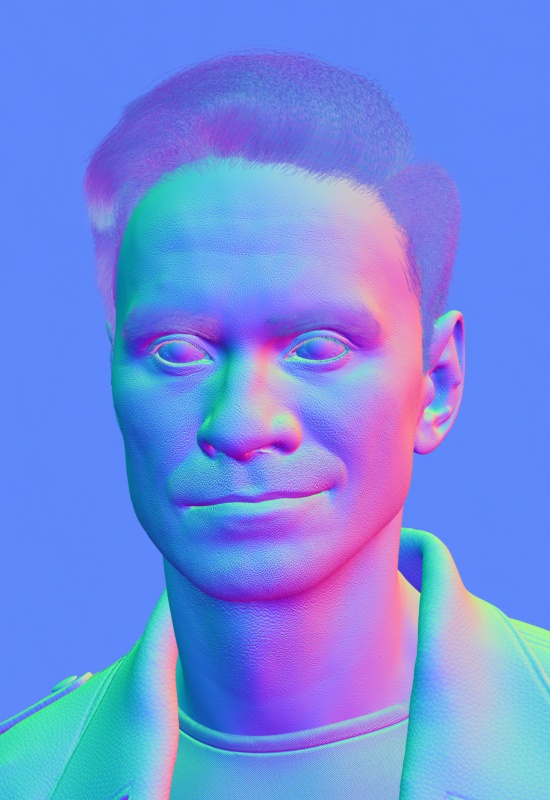}}%
    \insetwhite{Depth\vphantom{p}}{\includegraphics[width=\width]{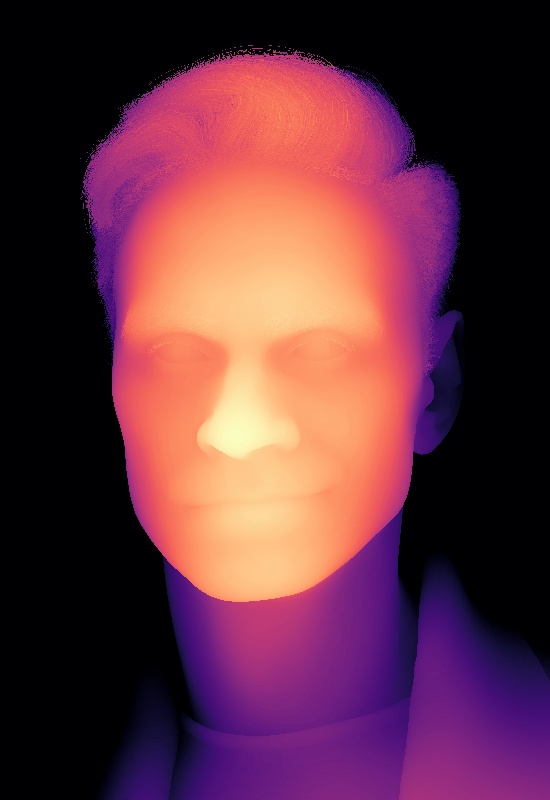}}%
    \insetwhite{Mask\vphantom{p}}{\includegraphics[width=\width]{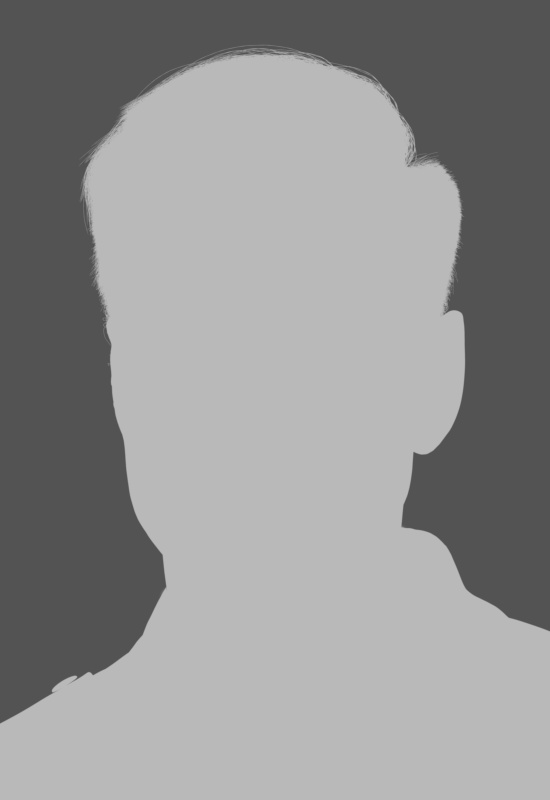}}%
    \insetwhite{UVs\vphantom{p}}{\includegraphics[width=\width]{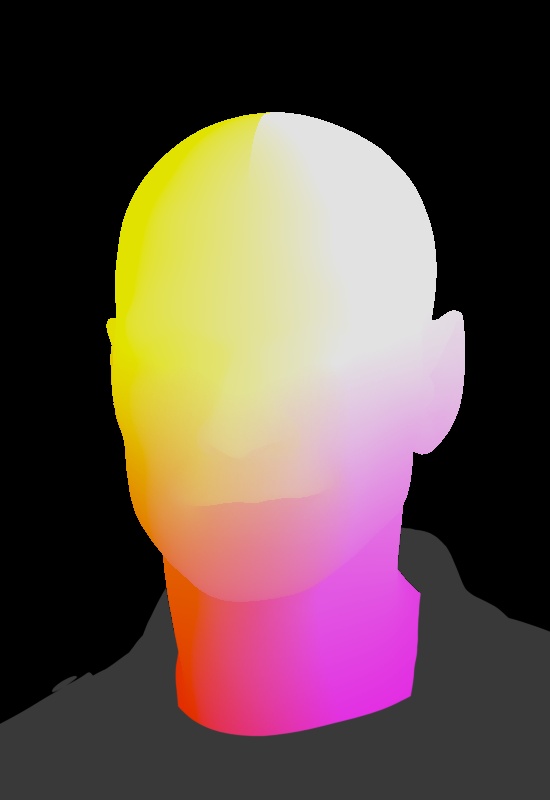}}%
    \insetwhite{Vertices\vphantom{p}}{\includegraphics[width=\width]{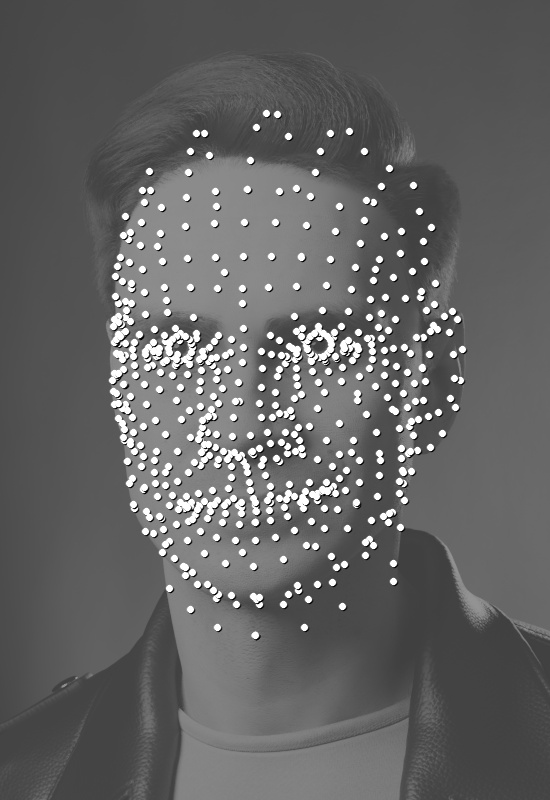}}

    \caption{We also synthesize labels for machine learning. Above are additional label types beyond those shown in \autoref{fig:intro_figure}.}
    \label{fig:gtlabels}
\end{figure}

In addition to rendering color images, we generate ground truth labels (see \autoref{fig:gtlabels}).
While our experiments in \autoref{sec:face_analysis} focus on landmark and segmentation annotations, synthetics lets us easily create a variety of rich and accurate labels
that enable new face-related tasks (see \autoref{sec:otherexamples}).

\section{Face analysis}
\label{sec:face_analysis}

We evaluate our synthetic data on two common face analysis tasks: face parsing and landmark localization.
We show that models trained on our synthetic data demonstrate competitive performance to the state of the art.
Note that all evaluations using our models are \textit{cross-dataset} -- we train purely on synthetic data and test on real data, while the state of the art evaluates \textit{within-dataset}, allowing the models to learn potential biases in the data. 

\newcommand{\imagerendertime}{five minutes}
\newcommand{\imageresolution}{512$\times$512}
\newcommand{\imagesamples}{256}
\newcommand{\datasetrendertime}{48 hours}
\newcommand{\datasetnumimgs}{100,000}

\subsection{Training methodology}
We render a \textbf{single} training dataset for both landmark localization and face parsing, comprising \datasetnumimgs{} images at \imageresolution{} resolution.
It took \datasetrendertime{} to render using 150 NVIDIA M60 GPUs.

During training, we perform data augmentation including
rotations, perspective warps, blurs, modulations to brightness and contrast, addition of noise, and conversion to grayscale. 
Such augmentations
are especially important for synthetic images which are otherwise free of imperfection (see \autoref{sec:ablation}).
While some of these could be done at render time, we perform them at training time in order to randomly apply different augmentations to the same training image.
We implemented neural networks with PyTorch~\cite{paszke2019pytorch}, and trained them with the Adam optimizer~\cite{kingma2014adam}.

\subsection{Face parsing}

Face parsing assigns a class label to each pixel in an image, e.g.\ skin, eyes, mouth, or nose.
We evaluate our synthetic training data on two face parsing datasets:
\textbf{Helen}~\cite{le2012interactive} is the best-known benchmark in the literature.
It contains 2,000 training images, 230 validation images, and 100 testing images,
each with 11 classes.
%
Due to labelling errors in the original dataset, we use Helen*~\cite{lin2019face}, a popular rectified version of the dataset which features corrected training labels, but leaves testing labels unmodified for a fair comparison.
\textbf{LaPa}~\cite{liu2020new} is a recently-released dataset which uses the same labels as Helen, but has more images, and exhibits more challenging expressions, poses, and occlusions.
It contains 18,176 training images, 2,000 validation images and 2,000 testing images.

As is common~\cite{lin2019face,liu2020new}, we use the provided 2D landmarks to align faces before processing.
We scale and crop each image so the landmarks are centered in a $512\!\times\!512$px region of interest.
Following prediction, we undo this transform to compute results
against the original label annotation, without any resizing or cropping.

\textbf{Method}
We treat face parsing as image-to-image translation.
Given an input color image $x$ containing $C$ classes, we wish to predict a $C$-channel label image $\hat{y}$ of the same spatial dimensions that matches the ground truth label image $y$.
Pixels in $y$ are one-hot encoded with the index of the true class.
For this, we use a UNet~\cite{ronneberger2015u} with ResNet-18 encoder~\cite{he2016resnet,Yakubovskiy:2019}.
We train this network with synthetic data only, minimizing a binary cross-entropy (BCE) loss between predicted and ground truth label images.
Note that there is nothing novel about our choice of architecture or loss function, this is a well-understood approach for this task.

\begin{figure}
    \centering
    \includegraphics[width=\linewidth]{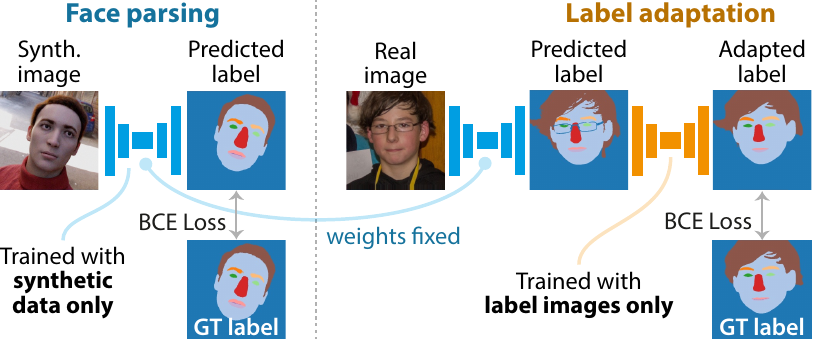}
    \caption{We train a face parsing network (using synthetic data only) followed by a label adaptation network to address systematic differences between synthetic and human-annotated labels.}
    \label{fig:face_parsing_arch}
\end{figure}

\begin{figure}
    \renewcommand\width{0.2\linewidth}
    \footnotesize
    \renewcommand{\arraystretch}{0.7}
    \begin{tabular}{@{}x{\width}@{}x{\width}@{}x{\width}@{}x{\width}@{}x{\width}@{}}
        Input & Trained with & + label & Trained with & Ground  \\
        (LaPa) &  synth. data & adaptation & real data & truth
    \end{tabular}

    \includegraphics[width=\linewidth]{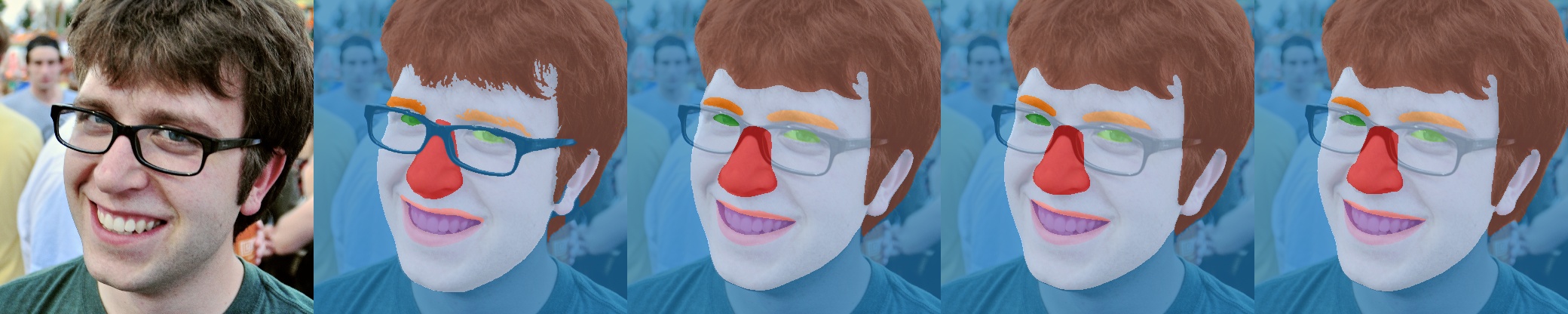}\vspace{0.1em}

    \includegraphics[width=\linewidth]{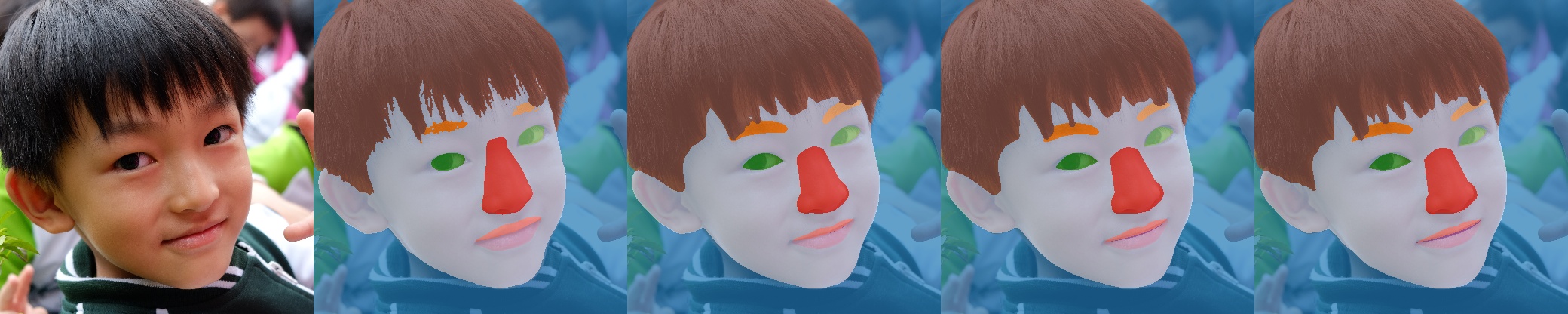}\vspace{0.1em}

    \includegraphics[width=\linewidth]{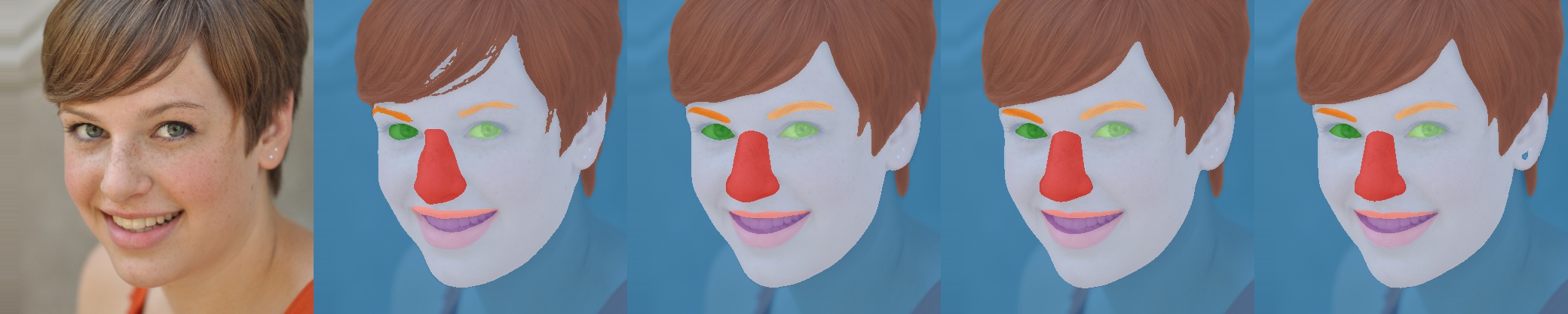}\vspace{0.1em}

    \caption{Face parsing results by networks trained with synthetic data (with and without label adaptation) and real data. Label adaptation addresses systematic differences between synthetic and real labels, e.g.\ the shape of the nose class, or granularity of hair.}
    \label{fig:face_parsing_qualitative_results}
\end{figure}

\textbf{Label adaptation.}
There are bound to be minor systematic differences between synthetic labels and human-annotated labels.
For example, where exactly is the boundary between the nose and the rest of the face?
To evaluate our synthetic data without needing to carefully tweak our synthetic label generation process for a specific real dataset, we use \textit{label adaptation}.
Label adaptation transforms labels predicted by our face parsing network (trained with synthetic data alone) into labels that are closer to the distribution in the real dataset (see \autoref{fig:face_parsing_arch}).
We treat label adaptation as another image-to-image translation task, and use a UNet with ResNet18 encoder~\cite{Yakubovskiy:2019}.
To ensure this stage is not able to `cheat', it is trained only on pairs of predicted labels $\hat{y}$ and ground truth labels $y$.
It is trained entirely separately from the face parsing network, and never sees any real images.

\textbf{Results} 
See Tables \ref{tab:helen} and \ref{tab:lapa} for comparisons against the state of the art, and \autoref{fig:face_parsing_qualitative_results} for some example predictions.
Although networks trained with our generic synthetic data do not outperform the state of the art, it is notable that they achieve similar results to previous work trained within-dataset on task-specific data.


\textbf{Comparison to real data.}
We also trained a network on the training portion of each real dataset to separate our training methodology from our synthetic data, presented as ``Ours (real)'' in Tables \ref{tab:helen} and \ref{tab:lapa}. 
It can be seen that training with synthetic data alone produces comparable results to training with real data.

\begin{table*}[t!]
    \centering
    \caption{A comparison with the state of the art on the Helen dataset, using F\textsubscript{1} score. As is common, scores for hair and other fine-grained categories are omitted to aid comparison to previous work. The overall score is computed by merging the nose, brows, eyes, and mouth categories. Training with our synthetic data achieves results in line with the state of the art, trained with real data.}
    \vspace{0.4em}

    \small
    \begin{tabular}{@{}l@{\hspace{3pt}}l c c c c c c c c c@{}}
        Method & & Skin & Nose & Upper lip & Inner mouth & Lower lip & Brows & Eyes & Mouth & Overall\\
        \hline
        \citet{guo2018residual} & \footnotesize{AAAI'18} & 93.8 & 94.1 & 75.8 & 83.7 & 83.1 & 80.4 & 87.1 & 92.4 & 90.5 \\
        \citet{wei2019accurate} & \footnotesize{TIP'19} & 95.6 & 95.2 & 80.0 & 86.7 & 86.4 & 82.6 & 89.0 & 93.6 & 91.6 \\
        \citet{lin2019face} & \footnotesize{CVPR'19} & 94.5 & 95.6 & 79.6 & 86.7 & 89.8 & 83.1 & 89.6 & 95.0 & 92.4 \\
        \citet{liu2020new} & \footnotesize{AAAI'20} & 94.9 & 95.8 & 83.7 & 89.1 & 91.4 & 83.5 & 89.8 & 96.1 & 93.1 \\
        \citet{te2020edge} & \footnotesize{ECCV'20} & 94.6 & 96.1 & 83.6 & 89.8 & 91.0 & 90.2 & 84.9 & 95.5 & 93.2 \\
        \hline
        Ours (real) & & 95.1 & 94.7 & 81.6 & 87.0 & 88.9 & 81.5 & 87.6 & 94.8 & 91.6 \\
        Ours (synthetic) & & 95.1 & 94.5 & 82.3 & 89.1 & 89.9 & 83.5 & 87.3 & 95.1 & 92.0 \\
    \end{tabular}
    
    \label{tab:helen}
\end{table*}

\begin{table*}[t!]
    \centering
    \caption{A comparison with the state of the art on LaPa, using F\textsubscript{1} score. For eyes and brows, L and R are left and right. For lips, U, I, and L are upper, inner, and lower. Training with our synthetic data achieves results in line with the state of the art, trained with real data.}
    \vspace{0.4em}

    \small
    \begin{tabular}{@{}l@{\hspace{3pt}}l c c c c c c c c c c c@{}}
        Method & & Skin & Hair & L-eye & R-eye & U-lip & I-mouth & L-lip & Nose & L-Brow & R-Brow & Mean \\
        \hline
        \citet{liu2020new} & \footnotesize{AAAI'20} & 97.2 & 96.3 & 88.1 & 88.0 & 84.4 & 87.6 & 85.7 & 95.5 & 87.7 & 87.6 & 89.8 \\
        \citet{te2020edge} & \footnotesize{ECCV'20} & 97.3 & 96.2 & 89.5 & 90.0 & 88.1 & 90.0 & 89.0 & 97.1 & 86.5 & 87.0 & 91.1 \\
        \hline
        Ours (real) & & 97.5 & 86.9 & 91.4 & 91.5 & 87.3 & 89.8 & 89.4 & 96.9 & 89.3 & 89.3 & 90.9 \\
        Ours (synthetic) & & 97.1 & 85.7 & 90.6 & 90.1 & 85.9 & 88.8 & 88.4 & 96.7 & 88.6 & 88.5 & 90.1 \\
        
    \end{tabular}
    
    \label{tab:lapa}
\end{table*}

\subsection{Landmark localization}
\label{sec:landmarks}


Landmark localization finds the position of facial points of interest in 2D.
We evaluate our approach on the \textbf{300W}~\cite{sagonas2016threew} dataset, which is split into common (554 images), challenging (135 images) and private (600 images) subsets.

\begin{figure}
    \renewcommand{\arraystretch}{0.5}
    \renewcommand\width{0.166\linewidth}
    \begin{tabular}{@{}c@{}c@{}c@{}c@{}c@{}c@{}}
        \includegraphics[width=\width]{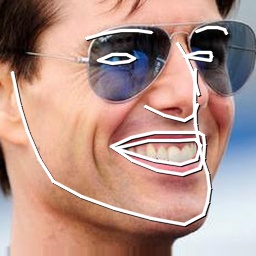}&%
        \includegraphics[width=\width]{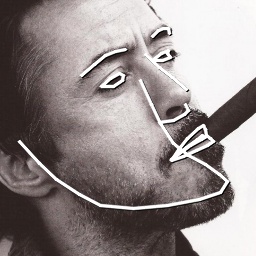}&%
        \includegraphics[width=\width]{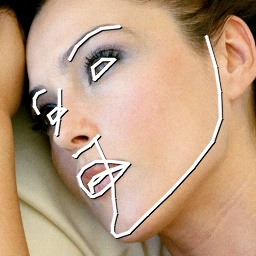}&%
        \includegraphics[width=\width]{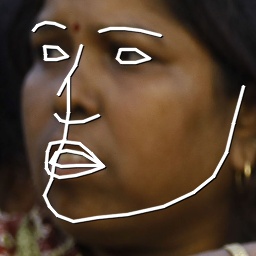}&%
        \includegraphics[width=\width]{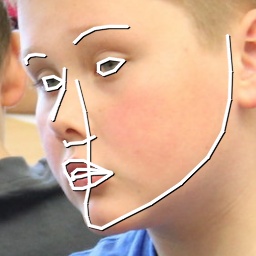}&%
        \includegraphics[width=\width]{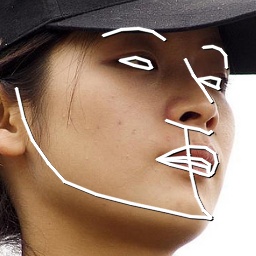}\\
        \includegraphics[width=\width]{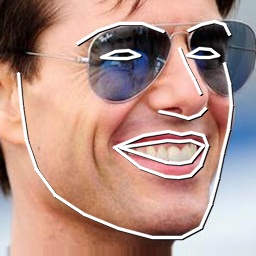}&%
        \includegraphics[width=\width]{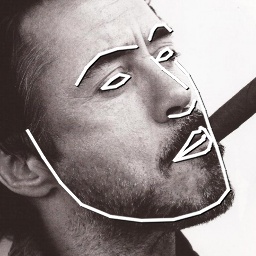}&%
        \includegraphics[width=\width]{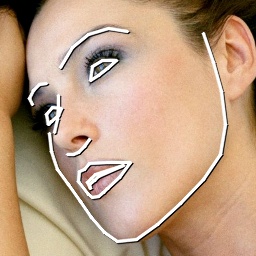}&%
        \includegraphics[width=\width]{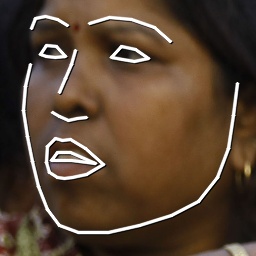}&%
        \includegraphics[width=\width]{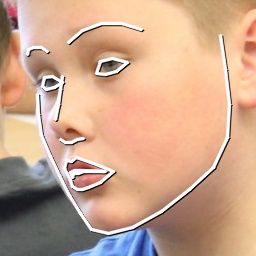}&%
        \includegraphics[width=\width]{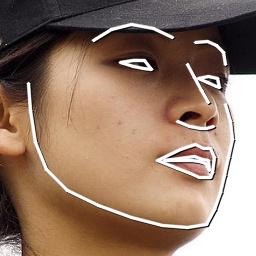}\\
    \end{tabular}\vspace{0.2em}

    \caption{Predictions before (top) and after (bottom) label adaptation. The main difference is changing the jawline from a 3D-to-2D projection to instead follow the facial outline in the image.}
    \label{fig:lmk_adapt}
\end{figure}

\textbf{Method}
We train a ResNet34~\cite{he2016resnet} with mean squared error loss to directly predict 68 2D landmark coordinates per-image.
We use the provided bounding boxes to extract a $256\!\times\!256$ pixel region-of-interest from each image.
The private set has no bounding boxes, so we use a tight crop around landmarks.

\begin{figure}
    \renewcommand{\arraystretch}{0.5}
    \renewcommand\width{0.166\linewidth}
    \begin{tabular}{@{}c@{}c@{}c@{}c@{}c@{}c@{}}
        \includegraphics[width=\width]{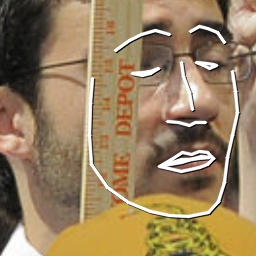}&%
        \includegraphics[width=\width]{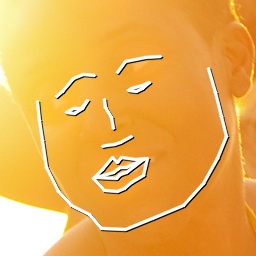}&%
        \includegraphics[width=\width]{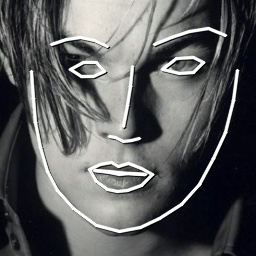}&%
        \includegraphics[width=\width]{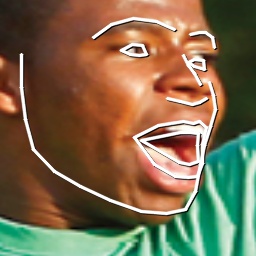}&%
        \includegraphics[width=\width]{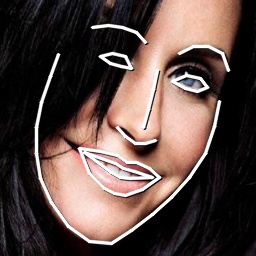}&%
        \includegraphics[width=\width]{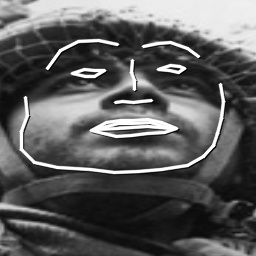}\\
        \includegraphics[width=\width]{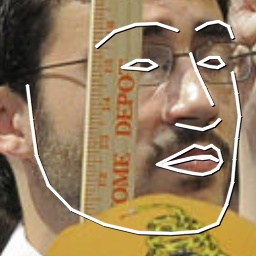}&%
        \includegraphics[width=\width]{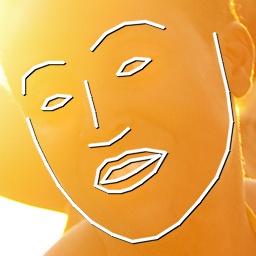}&%
        \includegraphics[width=\width]{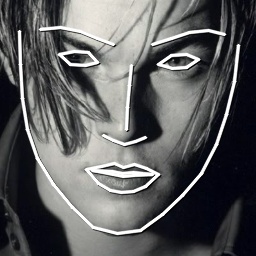}&%
        \includegraphics[width=\width]{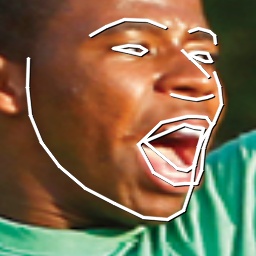}&%
        \includegraphics[width=\width]{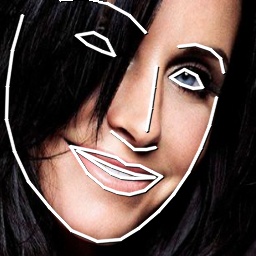}&%
        \includegraphics[width=\width]{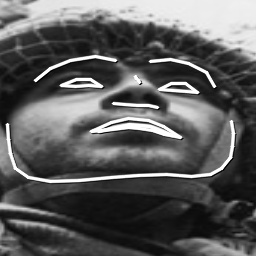}\\
    \end{tabular}\vspace{0.2em}
    
    \caption{Predictions by networks trained with real (top) and synthetic data (bottom). Note how the synthetic data network generalizes better across expression, illumination, pose, and occlusion.}
    \label{fig:lmk_real_vs_synth}
\end{figure}

\textbf{Label adaptation} is performed using a two-layer perceptron to address systematic differences between synthetic and real landmark labels (\autoref{fig:lmk_adapt}).
This network is never exposed to any real images during training.

\begin{table}[t!]
  \caption{Landmark localization results on the common, challenging, and private subsets of 300W. Lower is better in all cases. Note that 0.5 FR rate translates to 3 images, while 0.17 corresponds to 1.}\vspace{0.4em}
  \centering
  \small
  \begin{tabular}{@{}l@{\hspace{3pt}}l c c c@{}}
    & & Common & Challenging & Private \\ Method & & NME & NME & FR$_{10\%}$ \\
    \hline
    DenseReg~\cite{guler2017densereg} & {\footnotesize CVPR'17} & - & - & 3.67 \\
    LAB~\cite{wayne2018lab} & {\footnotesize CVPR'18} & 2.98 & 5.19 & 0.83 \\
    AWING~\cite{wang2019awing} & {\footnotesize ICCV'19} & 2.72 & 4.52 & 0.33\\
    ODN~\cite{zhu2019odn} & {\footnotesize CVPR'19} & 3.56 & 6.67 & - \\
    LaplaceKL~\cite{robinson2019laplace} & {\footnotesize ICCV'19} & 3.19 & 6.87 & - \\
    3FabRec~\cite{browatzki20203fabrec} & {\footnotesize CVPR'20}  & 3.36 & 5.74 & 0.17 \\
    \hline
    Ours (real) & & 3.37 & 5.77 & 1.17 \\
    Ours (synthetic) & & 3.09 & 4.86 & 0.50 \\[1.0ex]

    \multicolumn{5}{@{}l@{\hspace{3pt}}}{Ablation studies} \\
    \hline
    \multicolumn{2}{@{}l@{\hspace{3pt}}}{No augmentation} & 4.25 & 7.87 & 4.00 \\
    \multicolumn{2}{@{}l@{\hspace{3pt}}}{Appearance augmentation} & 3.93 & 6.80 & 1.83 \\
    \hline
    \multicolumn{2}{@{}l@{\hspace{3pt}}}{No hair or clothing} & 3.36 & 5.37 & 2.17 \\
    \multicolumn{2}{@{}l@{\hspace{3pt}}}{No clothing} & 3.20 & 5.09 & 1.00  \\
    \hline
    \multicolumn{2}{@{}l@{\hspace{3pt}}}{No label adaptation (synth.)} & 5.61 & 8.43 & 4.67\\
    \multicolumn{2}{@{}l@{\hspace{3pt}}}{No label adaptation (real)} & 3.44 & 5.71 & 1.17  \\
  \end{tabular}
  \label{tab:300W}
\end{table}

\textbf{Results} As evaluation metrics we use: Normalized Mean Error (NME)~\cite{sagonas2016threew} -- normalized by inter-ocular outer eye distance; and Failure Rate below a $10\%$ error threshold (FR$_{10\%}$).
See \autoref{tab:300W} for comparisons against state of the art on 300W dataset.
It is clear that the network trained with our synthetic data can detect landmarks with accuracy comparable to recent methods trained with real data.

\textbf{Comparison to real data}
We apply our training methodology (including data augmentations and label adaptation) to the the training and validation portions of the 300W dataset, to more directly compare real and synthetic data.
\autoref{tab:300W} clearly shows that training with synthetic data leads to better results, even when comparing to a model trained on real data and evaluated within-dataset.

\subsection{Ablation studies}
\label{sec:ablation}

\begin{figure}[t]
    \centering
    \includegraphics[width=1.0\linewidth]{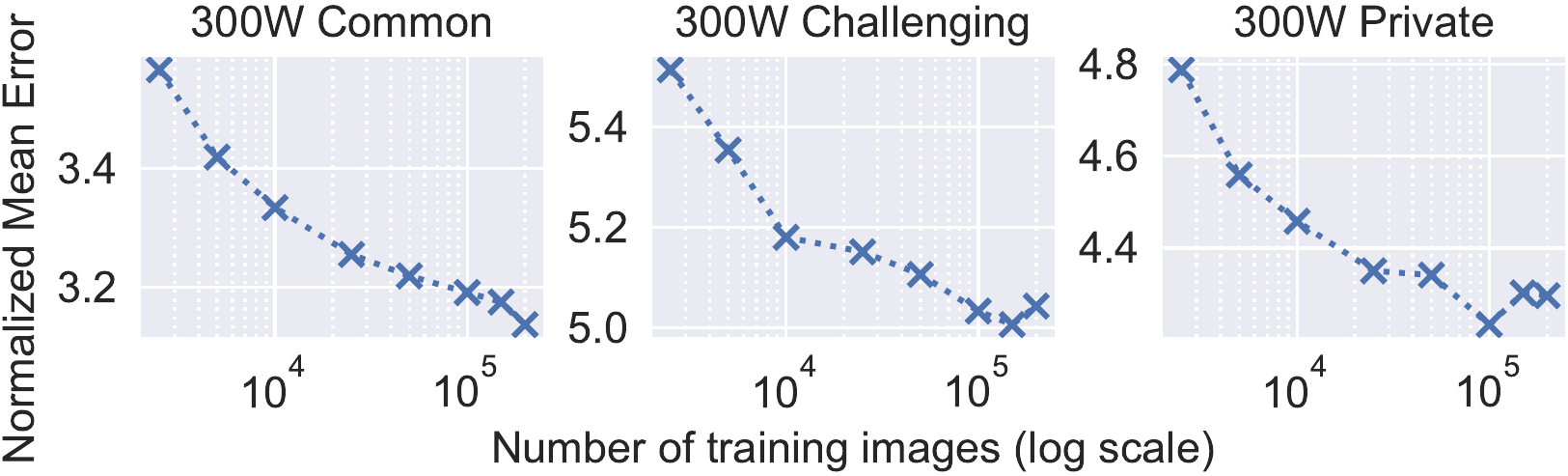}
    \caption{Landmark localization accuracy improves as we use more and more synthetic training data.}
    \label{fig:dataset_size}
\end{figure}

We investigate the effect of synthetic \textbf{dataset size} on landmark accuracy.
\autoref{fig:dataset_size} shows that landmark localization improves as we increase the number of training images, before starting to plateau at 100,000 images.

We study the importance of \textbf{data augmentation} when training models on synthetic data.
We train models with:
1) no augmentation;
2) appearance augmentation only (e.g.\ colour shifts, brightness and contrast);
3) full augmentation, varying both appearance and geometry (e.g.\ rotation and warping).
\autoref{tab:300W} shows the importance of augmentation, without which synthetic data does not outperform real.


\autoref{tab:300W} also shows the importance of \textbf{label adaptation} when evaluating models trained on synthetic data -- using label adaptation to improve label consistency reduces error.
Adding label adaptation to a model trained on real data results in little change in performance, showing that it does not benefit already-consistent within-dataset labels.


If we remove \textbf{clothing and hair}, landmark accuracy suffers (\autoref{tab:300W}).
This verifies the importance of our hair library and digital wardrobe, which improve the realism of our data.

Additional ablation studies analyzing the impact of render quality, and variation in pose, expression, and identity can be found in the supplementary material.

\subsection{Other examples}
\label{sec:otherexamples}

In addition to the quantitative results above, this section qualitatively demonstrates how we can solve additional problems using our synthetic face framework.

\begin{figure}
    \centering
    \includegraphics[width=0.495\linewidth]{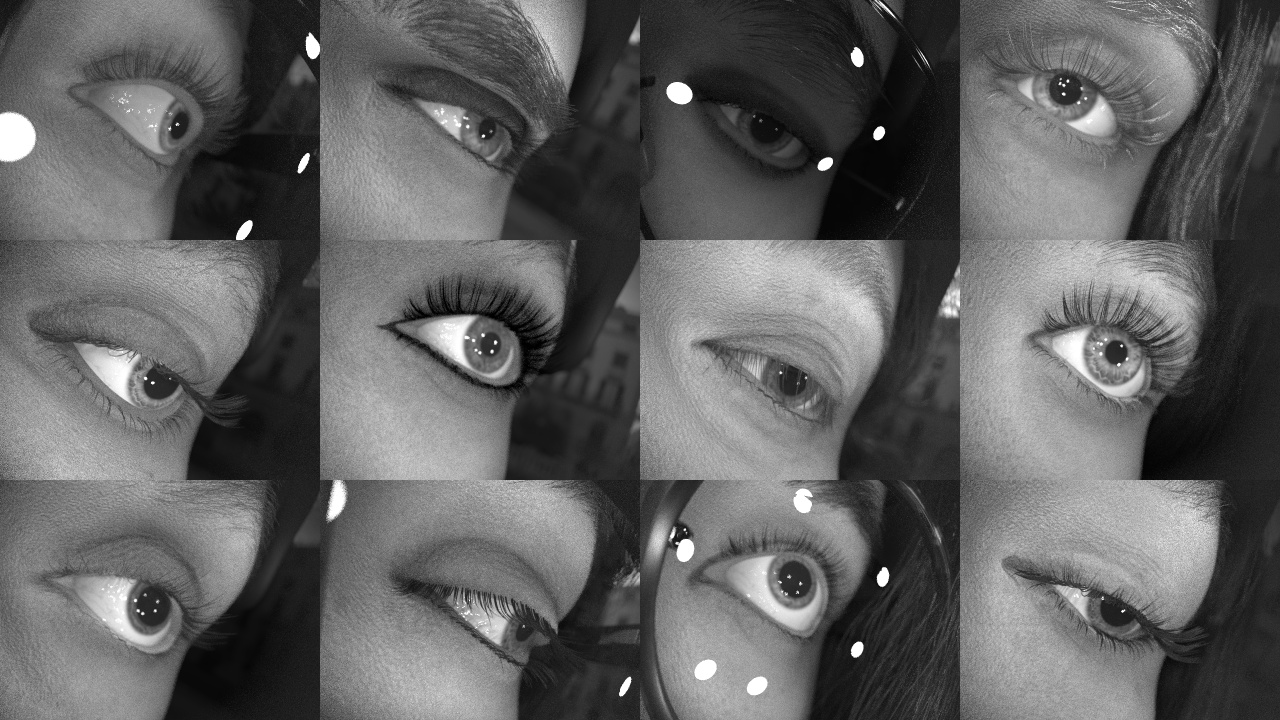}\hfill
    \includegraphics[width=0.495\linewidth]{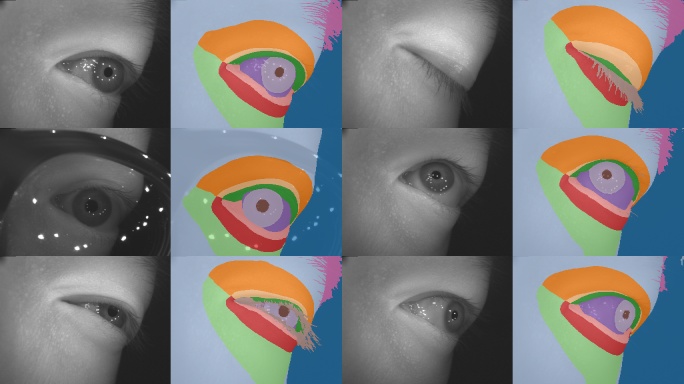}
    \caption{It is easy to generate synthetic training data for eye tracking (left) which generalizes well to real-world images (right).}
    \label{fig:eye_tracking}
\end{figure}

\textbf{Eye tracking} 
can be a key feature for virtual or augmented reality devices,
but real training data can be difficult to acquire~\cite{garbin2019openeds}.
Since our faces look realistic close-up, it is easy for us to set up a synthetic eye tracking camera and render diverse training images, along with ground truth.
\autoref{fig:eye_tracking} shows example synthetic training data for such a camera, along with results for semantic segmentation.

\begin{figure}
    \includegraphics[width=\linewidth]{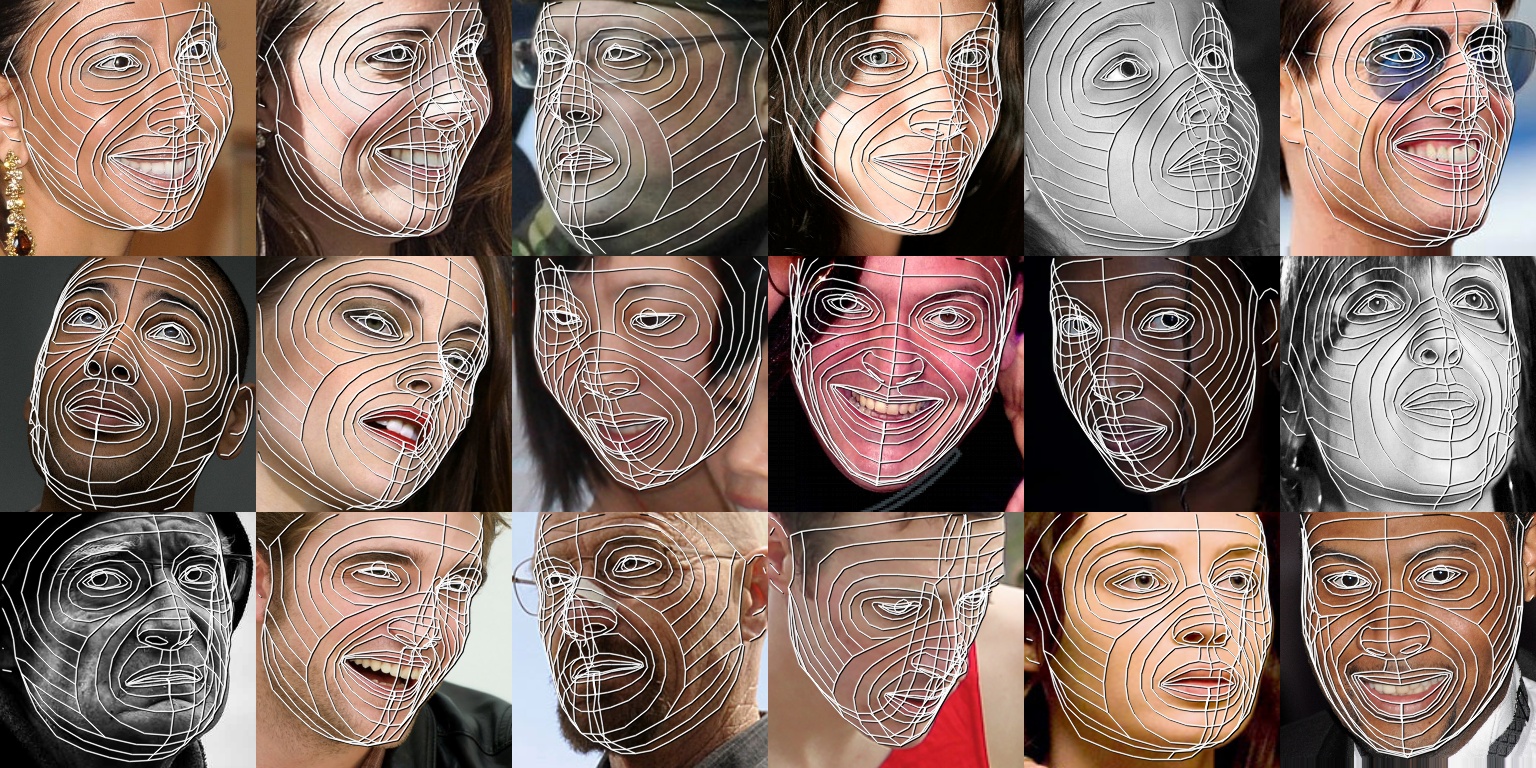}
    \caption{With synthetic data, we can easily train models that accurately predict ten times as many landmarks as usual. Here are some example dense landmark predictions on the 300W dataset.}
    \label{fig:dense}
\end{figure}

\textbf{Dense landmarks.} 
In \autoref{sec:landmarks}, we presented results for localizing 68 facial landmarks.
What if we wanted to predict ten times as many landmarks?
It would be impossible for a human to annotate this many landmarks consistently and correctly.
However, our approach lets us easily generate accurate dense landmark labels.
\autoref{fig:dense} shows the results of modifying our landmark network to regress 679 coordinates instead of 68, and training it with synthetic data.

\subsection{Discussion}

We have shown that it is possible to achieve results comparable with the state of the art for two well-trodden tasks: face parsing and landmark localization, without using a single real image during training.
This is important since it opens the door to many other face-related tasks that can be addressed using synthetic data in the place of real data.
%

Limitations remain.
As our parametric face model includes the head and neck only, we cannot simulate clothing with low necklines.
We do not include expression-dependent wrinkling effects, so realism suffers during certain expressions.
Since we sample parts of our model independently, we sometimes get unusual (but not impossible) combinations,
such as feminine faces that have a beard.
We plan to address these limitations with future work.


Photorealistic rendering is computationally expensive, so we must consider the environmental cost. 
In order to generate the dataset used in this paper, our GPU cluster used approximately 3,000kWh of electricity, equivalent to roughly 1.37 metric tonnes of CO\textsubscript{2}, 100\% of which was offset by our cloud computing provider.
This impact is mitigated by the ongoing progress of cloud computing providers to become carbon negative and use renewable energy sources~\cite{amazonco2, azureco2, googleco2}.
There is also the financial cost to consider.
Assuming \$1 per hour for an M60 GPU (average price across cloud providers), it would cost \$7,200 to render 100,000 images. 
Though this seems expensive, real data collection costs can run much higher, especially if we take annotation into consideration.



{\small
\textbf{Acknowledgements}
We thank Pedro Urbina, Jon Hanzelka, Rodney Brunet, and Panagiotis Giannakopoulos for their artistic contributions.
This work was published in ICCV 2021. The author list in the IEEE digital library is missing V.E. and M.J. due to a mistake on our side during the publishing process.}

{\small
\setlength{\bibsep}{0pt}
\bibliographystyle{abbrvnat}
\bibliography{00_main}
}

\end{document}